\documentclass[10pt]{article} 
\usepackage[accepted]{tmlr}

\usepackage{amsmath,amsfonts,bm}

\def\eqref#1{equation~\ref{#1}}

\def\1{\bm{1}}

\def\rvd{{\mathbf{d}}}

\def\rvw{{\mathbf{w}}}
\def\rvx{{\mathbf{x}}}

\def\rvz{{\mathbf{z}}}

\def\rmW{{\mathbf{W}}}
\def\rmX{{\mathbf{X}}}

\def\mZ{{\bm{Z}}}

\DeclareMathAlphabet{\mathsfit}{\encodingdefault}{\sfdefault}{m}{sl}
\SetMathAlphabet{\mathsfit}{bold}{\encodingdefault}{\sfdefault}{bx}{n}

\def\gL{{\mathcal{L}}}
\def\gM{{\mathcal{M}}}
\def\gN{{\mathcal{N}}}
\def\gO{{\mathcal{O}}}

\newcommand{\pdata}{p_{\rm{data}}}

\newcommand{\E}{\mathbb{E}}

\newcommand{\KL}{D_{\mathrm{KL}}}

\usepackage{hyperref}
\usepackage{url}
\usepackage{wrapfig}
\usepackage{graphicx}
\usepackage{caption}
\usepackage{subcaption}
\usepackage{booktabs}
\usepackage{amsmath}
\usepackage{multirow}
\usepackage{xcolor}
\usepackage{algorithm}
\usepackage{algpseudocode}

\definecolor{codebg}{RGB}{248,248,248}
\definecolor{codeframe}{RGB}{220,220,220}
\definecolor{pykw}{RGB}{0,0,160}
\definecolor{pystr}{RGB}{163,21,21}
\definecolor{pycmt}{RGB}{0,128,0}
\usepackage{listings}
\usepackage{inconsolata}   
\lstdefinestyle{pystyle}{
  language=Python,
  basicstyle=\ttfamily\small,
  keywordstyle=\color{pykw}\bfseries,
  stringstyle=\color{pystr},
  commentstyle=\color{pycmt}\itshape,
  showstringspaces=false,
  columns=fullflexible,
  numbers=left,
  numbersep=8pt,
  stepnumber=1,
  frame=single,
  rulecolor=\color{codeframe},
  backgroundcolor=\color{codebg},
  breaklines=true,
  tabsize=4,
  captionpos=b
}

\title{Learning Multimodal Energy-Based Model with Multimodal Variational Auto-Encoder via MCMC Revision}

\author{\name Jiali Cui \email jcui2@futurewei.com \\
      \addr Basking Ridge, New Jersey, USA\\
      Futurewei Technologies Inc
      \AND
      \name Zhiqiang Lao \email zlao@futurewei.com \\
      \addr Basking Ridge, New Jersey, USA\\
      Futurewei Technologies Inc
      \AND
      \name Heather Yu \email hyu@futurewei.com\\
      \addr Basking Ridge, New Jersey, USA\\
      Futurewei Technologies Inc
}

\def\month{05}  
\def\year{2026} 
\def\openreview{\url{https://openreview.net/forum?id=ZVD7bHNpY1}} 

\begin{document}

\maketitle

\begin{abstract}
Energy-based models (EBMs) are a flexible class of deep generative models and are well-suited to capture complex dependencies in \emph{multimodal} data. However, learning multimodal EBM by maximum likelihood requires Markov Chain Monte Carlo (MCMC) sampling in the joint data space, where noise-initialized Langevin dynamics often mixes poorly and fails to discover coherent \emph{inter-modal} relationships. Multimodal VAEs have made progress in capturing such inter-modal dependencies by introducing a shared latent generator and a joint inference model. However, both the shared latent generator and joint inference model are parameterized as unimodal Gaussian (or Laplace), which severely limits their ability to approximate the complex structure induced by multimodal data. In this work, we study the learning problem of the multimodal EBM, shared latent generator, and joint inference model. We present a learning framework that effectively interweaves their MLE updates with corresponding MCMC refinements in both the data and latent spaces. Specifically, the generator is learned to produce coherent multimodal samples that serve as strong initial states for EBM sampling, while the inference model is learned to provide informative latent initializations for generator posterior sampling. Together, these two models serve as complementary models that enable effective EBM sampling and learning, yielding realistic and coherent multimodal EBM samples. Extensive experiments demonstrate superior performance for multimodal synthesis quality and coherence compared to various baselines. We conduct various analyses and ablation studies to validate the effectiveness and scalability of the proposed multimodal framework.
\end{abstract}

\section{Introduction}
Deep generative models (DGMs) have achieved remarkable success in modelling complex data distributions for single modalities \citep{ho2020denoising, vahdat2020nvae, karras2020analyzing}. In recent years, these advances have rapidly extended to single-flow multimodal frameworks (e.g., text-to-image) \citep{ramesh2022hierarchical, alayrac2022flamingo, li2023blip} and to multi-flow multimodal models capable of supporting multiple generation flows within a single model \citep{Uni-d3, bao2023onefits,le2025one,li2025omniflow}. Among various DGMs, energy-based models (EBMs) form a particularly flexible class and are well suited to capture rich contextual structure in data space \citep{du2020improved, gao2020learning, Cui_2023_CVPR}. Yet, most existing EBMs focus on single-modality data and remain largely underexplored in the multimodal domain, falling behind other generative approaches.

Learning multimodal EBMs using maximum likelihood estimation (MLE) requires sampling from the EBM distribution, typically via Markov Chain Monte Carlo (MCMC) methods such as Langevin dynamics \cite{neal2011mcmc}. When the sampling chains are initialized from random noise, Langevin dynamics is often ineffective and may take a long time to mix between local modes \citep{nijkamp2020anatomy}. More critically, in multimodal modelling, effective EBM samples should additionally exhibit high coherence and consistency across modalities during the Langevin tranverse. Notably, to tackle EBM sampling challenge, various methods for \emph{single-modal} EBMs have been proposed, which can be broadly grouped into two categories: (i) using auxiliary models to amortize EBM sampling and thus bypass explicit MCMC during training, and (ii) using complementary models \citep{xie2018cooperative,xie2021learning,xie2022a,cui2023nips} to provide informative initial states, thereby enabling more effective EBM sampling. While both strategies can provide meaningful learning signals for single-modal EBMs, \emph{multimodal} data additionally contains complex inter-modal dependencies (i.e., relationships shared across modalities) and modality-specific variations (i.e., inductive biases unique to each modality). As a result, the development of multimodal EBM is needed yet remains largely underexplored.

Multimodal VAE \citep{wu2018multimodalmvae,shi2019variationalmmvae} has emerged as a promising approach for multimodal modelling. These models factorize a shared latent space and modality-specific generation models that map the low-dimensional shared latent space to the high-dimensional multimodal data space. The shared latent variables are thus learned to capture common structure across modalities, while the generators can preserve modality-specific characteristics. Learning typically follows the standard variational framework \citep{kingma2013auto}, using an inference model to approximate the generator posterior. Unlike the single-modality VAE, designing an effective joint inference model in the multimodal setting is nontrivial and remains an active research topic \citep{palumbo2023mmvae+,palumbo2024deepcmvae}. A prominent family of works adopts a mixture-of-experts (MoE) formulation \citep{shi2019variationalmmvae}, in which each modality contributes an expert posterior, typically parameterized as a unimodal Gaussian (or Laplace) over the shared latent variable. However, the multimodal generator posterior has a product-of-experts (PoE) structure that is often sharp and complex (see analysis in Section.~\ref{sec-revisit-mle}), making these MoE-based approximations inherently limited \citep{daunhawer2021limitations}. Alternatively to the variational learning scheme, one can sample from the generator posterior using MCMC methods such as Langevin dynamics \citep{han2017alternating}. Nevertheless, when chains are initialized from noise, Langevin dynamics again struggle to effectively explore the shared latent space \citep{short-run-multi-layer}, especially in complex multimodal regimes.

To address the learning and sampling challenges above, we propose a cooperative framework that incorporates the multimodal EBM, a shared latent generator, and a joint inference model into a unified probabilistic framework. We present a novel learning scheme that can seamlessly integrate these models by interweaving their MLE updates. Specifically, to enable effective multimodal EBM sampling, the shared latent generator is learned to approximate the EBM distribution so that it can provide informative and coherent initial states for MCMC-based EBM sampling. For the mismatch between the MoE-based joint inference model and the PoE-style generator posterior, we perform a finite number of MCMC posterior update steps, initialized from the joint inference model, which is itself learned to approximate the generator posterior. This cooperative interplay among the three models yields effective sampling, accurate posterior inference, and stable training dynamics. Various experiments demonstrate that our learned EBM and shared latent generator model produce realistic and highly coherent multimodal samples, and consistently outperform strong multimodal variational baselines in both synthesis quality and cross-modal coherence. We further provide ablations and diagnostic analyses to clarify how each component contributes to sampling efficiency, multimodal coherence, and overall performance.

Our contributions can be summarized as: (\textbf{i}) We present a novel learning methodology that facilitates effective EBM sampling and learning toward multimodality. (\textbf{ii}) We integrate the multimodal EBM, shared latent generator, and joint inference model into a unified probabilistic framework, interleaving their MLE updates so that each component benefits from the others. (\textbf{iii}) We conduct extensive experiments, demonstrating superior performance of our multimodal EBM and effectiveness of our learning method.

\section{Preliminary}
In this section, we briefly review energy-based models, shared latent generator models for multimodal data, and joint inference, which together form the foundation of our framework.

\subsection{Preliminary: Multimodal Energy-based Model}
Let $\rmX = \{\rvx_1, \dots, \rvx_M\}$ denote an observed multimodal data example consisting of $M$ modalities, and let $\pdata(\rmX)$ represent the unknown empricial data distribution. Energy-based models (EBMs) \citep{du2020improved,gao2020learning,Cui_2023_ICCV} represent a flexible class of generative models that define an undirected probability distribution
\begin{equation}
    \pi_\alpha(\rmX)=\frac{1}{\mZ(\alpha)}\exp\left[F_\alpha(\rmX)\right]
\end{equation}
where $\mZ(\alpha) = \int \exp\big[F_\alpha(\rmX)\big] \rvd\rmX$ is the normalizing constant (or partition function), and $F_\alpha(\rmX)$ is a scalar energy function parameterized by $\alpha$. For multimodal $\rmX$, the energy function takes all inputs $\{\rvx_1, \dots, \rvx_M\}$ and outputs a single energy value.

While EBMs offer considerable modelling flexibility, their application to multimodal data remains relatively underexplored. A central challenge is to design energy functions that effectively capture the joint structure and dependencies across heterogeneous modalities. In this work, we adopt a simple yet general parameterization: $F_\alpha(\rmX) = \bar{f}([f_1(\rvx_1), \dots, f_M(\rvx_M)])$, where each $f_i$ maps modality $\rvx_i$ to a fixed-dimensional feature vector, and $\bar{f}$ aggregates the concatenated features to produce the final energy score. More sophisticated architectures could further improve performance, but our focus is on the learning methodology rather than architectural innovation. Implementation details are provided in Appendix.~\ref{sec-app-implement}.

\noindent\textbf{EBM Learning and Sampling.} Given $N$ multimodal data $\{\rmX_1, \dots, \rmX_N\}$ drawn from $\pdata(\rmX)$, the EBM can be learned via maximum likelihood estimation (MLE). The log-likelihood is computed as $\gL_\pi(\alpha)=\frac{1}{N}\sum_{i=1}^{N}\log \pi_\alpha(\rmX_i)$. When $N$ becomes sufficiently large, maximizing $\gL_\pi(\alpha)$ is equivalent to minimizing the KL divergence between the true data density and EBM density, i.e., 
\begin{equation}\label{eq-ebm-mle-grad}
    -\gL_\pi(\alpha) = \KL(\pdata(\rmX)||\pi_\alpha(\rmX))\;\;\;\text{where}
\end{equation}
\begin{equation}
    \frac{\partial}{\partial \alpha}\gL_\pi(\alpha) = \E_{\pdata(\rmX)}\left[\frac{\partial}{\partial \alpha}F_\alpha(\rmX)\right] - \E_{\pi_\alpha(\rmX)}\left[\frac{\partial}{\partial \alpha}F_\alpha(\rmX)\right]\nonumber
\end{equation}
Computing Eqn.~\ref{eq-ebm-mle-grad} requires EBM samples, i.e., $\rmX \sim \pi_\alpha(\rmX)$, which can be achieved via MCMC methods, such as Langevin dynamics \citep{neal2011mcmc} that iteratively updates
\begin{equation}\label{eq-ebm-ld}
    \rmX^{k+1}=\rmX^{k}+s\frac{\partial}{\partial \rmX^{k}}\log \pi_\alpha(\rmX^{k}) + \sqrt{2s}\cdot\epsilon^{k}
\end{equation}
where $k$ denotes the iteration index, $s$ is the step size, and $\epsilon^{k}$ is Gaussian noise. In the limit as $s \rightarrow 0$ and $k \rightarrow \infty$, this process will converge to the stationary distribution $\pi_\alpha(\rmX)$ \citep{neal2011mcmc}. 

\noindent\textbf{Multimodal Challenge.} Common practices typically adopt short-run Langevin dynamics \citep{nijkamp2019learning}, which perform certain steps (e.g., $k=30$) of Langevin dynamics to obtain approximate EBM samples. Although this yields useful learning signals, it is still difficult to obtain high-quality samples when the chain is initialized from non-informative states\footnote{For example, $\rmX^{k=0}$ drawn from a unit Gaussian or a uniform distribution.} \citep{grathwohl2021no,kumar2019maximum}.

For multimodal data, this difficulty is amplified. The input $\rmX = \{\rvx_1, \dots, \rvx_M\}$ exhibits complex inter-modal dependencies (i.e., relationships among modalities), and successful sampling should traverse coherent local modes that preserve consistent cross-modal structure. As a result, informative and \emph{coherent} initializations become especially critical for effective multimodal EBM sampling and learning.

\subsection{Preliminary: Multimodal Variational AutoEncoder}
To capture inter-modal relationships, shared latent generator models have emerged as a promising approach for multimodal modelling \citep{wu2018multimodalmvae,shi2019variationalmmvae}. Let $\rvz$ denote a low-dimensional latent variable. The shared latent generator specifies a joint distribution over multimodal inputs as
\begin{equation}\label{eq-shared-gen}
\begin{aligned}
    &p_\omega(\rmX, \rvz)=p_\omega(\rmX|\rvz)p_0(\rvz)\;\;\; \text{where}\\
    p_\omega(\rmX|\rvz)&=p_{\omega_1}(\rvx_1|\rvz)p_{\omega_2}(\rvx_2|\rvz)\cdots p_{\omega_M}(\rvx_M|\rvz)
\end{aligned}
\end{equation}
Here, $p_0(\rvz)$ is the prior distribution (e.g., Gaussian or Laplace distribution) over a shared latent variable $\rvz$, and $p_\omega(\rmX|\rvz)$ is the conditional likelihood given such shared latent variable and factorizes a product of $M$ modality-specific generation models. Each $p_{\omega_i}(\rvx_i|\rvz) \sim \gN(\mu_\omega(\rvz), \sigma^2 I_d)$ represents a conditional Gaussian parameterized by $\omega_i$, mapping the low-dimensional latent space to high-dimensional data space.

This shared latent generator model is designed to capture modality-invariant representations (i.e., high-level semantics) across different modalities through the shared latent space $\rvz$, while also being capable of modelling modality-specific biases through separate generation models for each modality. 

\noindent\textbf{Multimodal Joint Inference Model}. For learning Eqn.~\ref{eq-shared-gen}, multimodal VAEs \citep{sutter2020multimodalmmJSD,hwang2021multiMVTCAE,palumbo2023mmvae+,palumbo2024deepcmvae} employ variational learning schemes by introducing a joint inference model. For multimodal data, factorizing effective joint inference models remains challenging and is an active research area (see details in Section.~\ref{sec-related-work}). Among various methods, one major paradigm is the mixture-of-experts (MoE) \citep{shi2019variationalmmvae} defined as
\begin{equation}\label{eq-inf}
    q_\phi(\rvz|\rmX) = \frac{1}{M}\sum_{i=1}^M q_{\phi_i}(\rvz|\rvx_i)
\end{equation}
Each $q_{\phi_i}(\rvz|\rvx_i) \sim \gN(\mu_{\phi_i}(\rvx_i), V_{\phi_i}(\rvx_i))$ is modeled as conditional Gaussian (or Laplace), where $\mu_{\phi_i}(\rvx_i)$ and $V_{\phi_i}(\rvx_i)$ denote the mean and diagonal covariance matrix parameterized by $\phi_i$. 

This mixture-based joint inference model offers a tractable approximation to the generator posterior, particularly useful in scenarios with missing modalities. However, both the mixture formulation and the assumed individual posteriors are limited in statistical expressivity. They often induce an overly smooth latent space, which may fail to capture the intricate structure of the true multimodal generator posterior, ultimately resulting in a suboptimal generator model.

\subsubsection{Revisiting Learning from MLE Perspective}\label{sec-revisit-mle}
In this section, we revisit the learning of the shared latent generator from the maximum-likelihood perspective. Our goal is to clarify the use of MCMC-based revision and a joint inference model in our framework.

\noindent\textbf{MLE Learning:} Consider maximizing the log-likelihood of the shared latent generator model, i.e., $\gL_p(\omega)=\frac{1}{N}\sum_{i=1}^{N}\log p_\omega(\rmX_i)$, where $p_\omega(\rmX_i)=\int_\rvz p_\omega(\rmX, \rvz) \rvd\rvz$ is its marginal distribution. With a sufficiently large number of $N$, it is equivalent to minimizing the KL-divergence as
\begin{equation}\label{eq-gen-mle-grad}
    \gL_p(\omega) = \KL(\pdata(\rmX)||p_\omega(\rmX))
\end{equation}
\begin{equation}
    \text{where}\;\;\;\frac{\partial}{\partial \omega}\gL_p(\omega) = \E_{\pdata(\rmX)p_\omega(\rvz|\rmX)}\left[\frac{\partial}{\partial \omega}\log p_\omega(\rmX, \rvz)\right]\nonumber
\end{equation}
Here, $p_\omega(\rvz|\rmX)$ is the generator posterior. To shed further light, we can decompose it into
\begin{equation}\label{eq-gen-posterior}
    p_\omega(\rvz|\rmX) = \frac{p_\omega(\rmX|\rvz)p_0(\rvz)}{p_\omega(\rmX)}=\frac{p_0(\rvz)}{p_\omega(\rmX)}\prod_{i=1}^M\frac{p_{\omega_i}(\rvz|\rvx_i)p_{\omega_i}(\rvx_i)}{p_0(\rvz)}\propto\frac{\prod_{i=1}^Mp_{\omega_i}(\rvz|\rvx_i)}{\prod_{i=1}^{M-1}p_0(\rvz)}
\end{equation}
which reveals that the generator posterior is effectively a product of individual posteriors, modulated by the prior, leading to sharp and complex structures in the latent space. Approximating this product-based posterior using a mixture-based joint inference model (Eqn.~\ref{eq-inf}) can be suboptimal due to the smoothing effect inherent in averaging \citep{daunhawer2021limitations}. Moreover, the unimodal $q_{\phi_i}(\rvz|\rvx_i)$ (e.g., Gaussian or Laplace) can be limited in statistical expressivity and may fail to capture the intricate structure of the complex individual posteriors \citep{pang2021generative, xie2022a}.

\noindent\textbf{MCMC Posterior Sampling.} Alternatively, one can obtain posterior samples by MCMC methods, such as Langevin dynamics \citep{han2017alternating,deqian-short-run-molecule,deqian-short-run-plan}, i.e., 
\begin{equation}\label{eq-gen-ld}
    \rvz^{k+1}=\rvz^{k}+s\frac{\partial}{\partial \rvz^{k}}\log p_\omega(\rvz^{k}|\rmX) + \sqrt{2s}\cdot\epsilon^{k}
\end{equation}
The target distribution is the generator posterior $p_\omega(\rvz|\rmX)$, and the gradient term can be computed as $\frac{\partial}{\partial \rvz}\log p_\omega(\rvz|\rmX) \propto \frac{\partial}{\partial \rvz}\log p_\omega(\rmX|\rvz)p_0(\rvz)$. The log-likelihood gradient decomposes as $\log p_\omega(\rmX|\rvz) = \sum_{i=1}^M \log p_{\omega_i}(\rvx_i|\rvz)$, which updates shared latent variable $\rvz$ to explain all modality observations $\{\rvx_1, \dots, \rvx_M\}$. As $s \rightarrow 0$ and $k \rightarrow \infty$, this process converges to the stationary distribution $p_\omega(\rvz|\rmX)$ \citep{neal2011mcmc}.

\noindent\textbf{Multimodal Challenge.} While MCMC-based posterior sampling can yield more accurate approximations than variational methods, it often suffers from poor mixing and slow convergence when using short-run Langevin dynamics (e.g., $k=10$) initialized from non-informative points\footnote{e.g., $\rvz^{k=0}$ drawn from a unit Gaussian or uniform distribution.} \citep{short-run-multi-layer}. More critically, the product-based formulation of the generator posterior requires complete observations from all modalities, and the individual unimodal posteriors are often undertrained and poorly calibrated. With inconsistent initializations, this becomes ill-defined in cross-modal inference scenarios, where only a subset of modalities is available \citep{shi2019variationalmmvae,daunhawer2021limitations}.

\section{Multimodal Learning via MCMC Revision}
To address the learning and sampling challenges discussed above, we propose a multimodal learning framework that jointly learns the multimodal EBM, the shared latent generator, and the joint inference model through a cooperative mechanism. The key idea is to leverage the \textit{complementary} strengths of each component: the joint inference model provides informative and coherent initial states for MCMC posterior sampling; the shared latent generator produces consistent multimodal samples that initialize EBM sampling; and the EBM, in turn, supplies revision signals that refine both the generator and the inference model. By interweaving their maximum-likelihood updates with MCMC-based revision, each model benefits from the others, which improves sampling effectiveness and leads to more effective multimodal modelling.

\subsection{Revision Signal of MCMC Kernels}
Let $\gM_{\alpha}^{k_\rmX}(\cdot)$ denote the Markov transition kernel corresponding to $k_\rmX$ steps of Langevin dynamics on EBM density (Eqn.~\ref{eq-ebm-ld}), and let $\gM_{\omega}^{k_\rvz}(\cdot)$ denote the Markov transition kernel for $k_\rvz$ steps of Langevin dynamics on generator posterior (Eqn.~\ref{eq-gen-ld}). We define two joint densities that describe these MCMC revision processes:
\begin{equation}\label{eq-joint-density}
\begin{aligned}
        &\Omega_{\omega, \alpha}(\rmX,\rvz)=\gM_{\alpha}^{k_\rmX}\cdot p_\omega(\rmX|\rvz)p_0(\rvz)\\
        &\Phi_{\omega,\phi}(\rmX, \rvz)=\pdata(\rmX)\cdot\gM_{\omega}^{k_\rvz}\cdot q_\phi(\rvz|\rmX)
\end{aligned}
\end{equation}
where $\Omega_{\omega, \alpha}(\rmX,\rvz)$ is obtained by first drawing $\rvz \sim p_0(\rvz)$, generating an initial multimodal sample $\rmX \sim p_\omega(\rmX|\rvz)$, and then applying $k_\rmX$ steps of Langevin dynamics under the EBM to obtain $\rmX$. The resulting marginal $\gM_{\alpha}^{k_\rmX}p_\omega(\rmX)=\int_{\bar{\rmX}}\int_\rvz\gM_{\alpha}^{k_\rmX}(\bar{\rmX})p_\omega(\bar{\rmX}, \rvz)\rvd\rvz \rvd\bar{\rmX}$ is generally more expressive than the Gaussian (or Laplace) generator. Similarly, $\Phi_{\omega,\phi}(\rmX, \rvz)$ is obtained by first taking a data example $\rmX \sim \pdata(\rmX)$, initializing the latent state via the joint inference model $\rvz \sim q_\phi(\rvz|\rmX)$, and then applying $k_\rvz$ steps of Langevin dynamics targeting the generator posterior. This yields a revised posterior marginal $\gM_{\omega}^{k_\rvz}q_\phi(\rvz|\rmX)=\int_{\bar{\rvz}}\gM_{\omega}^{k_\rvz}(\bar{\rvz}) q_\phi(\bar{\rvz}|\rmX)\rvd\bar{\rvz}$, which can better approximate the sharp, PoE structure of the multimodal generator posterior than the mixture-based approximate posterior and its unimodal (Gaussian or Laplace) experts.

Related MCMC-revised densities have been used in earlier work \citep{xie2018cooperative,xie2022a,xie2021learning,cui2023nips} on single-modality EBMs, where a generator provides initial states that are refined by an EBM, and the resulting samples are then used to update both models. In contrast, our formulation operates directly on multimodal data $\rmX$ and the MCMC kernels act on both the EBM and the multimodal generator posterior. In particular, $\Omega_{\omega, \alpha}(\rmX,\rvz)$ leverages the shared latent generator to produce coherent multimodal initializations (across all $\rvx_i$), which are then refined by the EBM. This improves mixing and learning for multimodal EBM sampling while respecting inter-modal dependencies. $\Phi_{\omega,\phi}(\rmX, \rvz)$ uses the mixture-based joint inference model as an amortized initializer for latent MCMC. The subsequent Langevin refinement yields more accurate posterior samples than those obtained by purely variational multimodal VAE schemes \citep{shi2019variationalmmvae,palumbo2023mmvae+}, especially in regimes where the true posterior is sharp and highly structured.
These two revised joint densities provide the core “revision signals” in our framework: the EBM refines generator samples in data space, and the generator posterior refines inference samples in latent space. In Section.~\ref{sec-learning}, we show how to use $\Omega_{\omega, \alpha}(\rmX,\rvz)$ and $\Phi_{\omega,\phi}(\rmX, \rvz)$ to define coupled learning objectives that jointly update all three models.\\

\subsection{Learning Objectives with MCMC-revised Densities} \label{sec-learning}
The two MCMC-revised joint densities serve as intermediate targets that couple the three models. At optimization step $t$, MCMC kernels are evaluated at the current parameters for revised joint densities denoted as $\Omega_{\omega_t, \alpha_t}(\rmX, \rvz)$ and $\Phi_{\omega_t, \phi_t}(\rmX, \rvz)$. During parameter updates, we adopt a stop-gradient operation for the revision process, so that the revised samples are treated as fixed guidance, and gradients are not propagated through the Langevin transition kernels. Each model is then updated by minimizing a KL divergence between these revised densities and its own current distribution. In this way, the EBM, the shared latent generator, and the joint inference model are all guided toward the more accurate “revised” distributions.

(\textbf{i}) for multimodal EBM $\alpha$, we define the learning objective $\gL_\pi(\alpha)$ as
\begin{equation}\label{eq-ebm-grad}
    - \gL_\pi(\alpha) = \KL(\Phi_{\omega_t,\phi_t}(\rmX, \rvz)||\pi_\alpha(\rmX)q_\phi(\rvz|\rmX)) -  \KL(\Omega_{\omega_t, \alpha_t}(\rmX,\rvz)||\pi_\alpha(\rmX)q_\phi(\rvz|\rmX))
\end{equation}
\begin{equation}
\begin{aligned}
\text{with}\;\;\;& \frac{\partial}{\partial \alpha}\gL_\pi(\alpha) 
    = \E_{\Phi_{\omega_t, \alpha_t}(\rmX,\rvz)}\left[\frac{\partial}{\partial \alpha}F_\alpha(\rmX)\right] - \E_{\Omega_{\omega_t, \alpha_t}(\rmX,\rvz)}\left[\frac{\partial}{\partial \alpha}F_\alpha(\rmX)\right]\nonumber
\end{aligned}
\end{equation}
Intuitively, our multimodal EBM is updated to increase the energy (i.e., lower probability) of samples from $\Omega$ to those from $\Phi$. Since $\Phi$ is anchored on real data and refined in latent space, while $\Omega$ is obtained from generator-initialized chains refined in data space, this contrast encourages the EBM to better align with the data-driven revised distribution. We optimize $\gL_\pi(\alpha)$ via stochastic gradient ascent (SGA). 

(\textbf{ii}) For shared latent generator model $\omega$, the learning objective $\gL_p(\omega)$ is
\begin{equation}\label{eq-gen-grad}
    - \gL_p(\omega) = \KL(\Phi_{\omega_t,\phi_t}(\rmX, \rvz)||p_\omega(\rmX, \rvz)) +  \KL(\Omega_{\omega_t, \alpha_t}(\rmX,\rvz)||p_\omega(\rmX, \rvz))
\end{equation}
\begin{equation}
\begin{aligned}
\text{with}\;\;\;& \frac{\partial}{\partial \omega}\gL_p(\omega) 
    = \E_{\Phi_{\omega_t, \alpha_t}(\rmX,\rvz)}\left[\frac{\partial}{\partial \omega}\log p_\omega(\rmX, \rvz)\right] - \E_{\Omega_{\omega_t, \alpha_t}(\rmX,\rvz)}\left[\frac{\partial}{\partial \omega}\log p_\omega(\rmX, \rvz)\right]\nonumber
\end{aligned}
\end{equation}
The shared latent generator is learned toward both revised distributions: $\Phi$, which reflects data-driven posterior refinement, and $\Omega$, which reflects EBM-refined samples. Matching these two targets guides the generator to better approximate the EBM density while maintaining consistency with data and posterior structure. We again use SGA to update $\omega$.

(\textbf{iii}) For multimodal joint inference model $\phi$, the learning objecitve $\gL_q(\phi)$ is 
\begin{equation}\label{eq-inf-grad}
    - \gL_q(\phi) = \KL(\Phi_{\omega_t,\phi_t}(\rmX, \rvz)||\pdata(\rmX)q_\phi(\rvz|\rmX)) +  \KL(\Omega_{\omega_t, \alpha_t}(\rmX,\rvz)||\pi_\alpha(\rmX)q_\phi(\rvz|\rmX))
\end{equation}
\begin{equation}
\begin{aligned}
    \text{with}\;\;\;& \frac{\partial}{\partial \phi}\gL_q(\phi) 
    = \E_{\Phi_{\omega_t, \alpha_t}(\rmX,\rvz)}\left[\frac{\partial}{\partial \phi}\log q_\phi(\rvz|\rmX)\right] + \E_{\Omega_{\omega_t, \alpha_t}(\rmX,\rvz)}\left[\frac{\partial}{\partial \phi}\log q_\phi(\rvz|\rmX)\right]\\
    &=\E_{\Phi_{\omega_t, \alpha_t}(\rmX,\rvz)}\left[\frac{\partial}{\partial \phi}\text{$\mathrm{LSE}$}(\log q_{\phi_i}(\rvz|\rvx_i))\right] + \E_{\Omega_{\omega_t, \alpha_t}(\rmX,\rvz)}\left[\frac{\partial}{\partial \phi}\text{$\mathrm{LSE}$}(\log q_{\phi_i}(\rvz|\rvx_i))\right]\nonumber
\end{aligned}
\end{equation}
where $\mathrm{LSE}(\cdot)$ denotes the log-sum-exp operation, i.e., $\log \sum_{i=1}^M \exp(\cdot)$, corresponding to the mixture-based joint inference over $M$ modalities. In this case, the inference model is encouraged to align with both revised joint densities: one anchored on the true data distribution $\pdata$ and one on the current EBM $\pi_\alpha$. Because the mixture-of-experts structure is preserved through the log-sum-exp form, the gradients naturally propagate to each modality-specific encoder. We update $\phi$ using SGA based on the computed gradient.


\subsubsection{Why MCMC-revised Kernels?}
The MCMC-revised kernels act as bridges between the three components and determine how they interact during cooperative learning. Consider the long-run behavior of the transition kernels $\gM_{\alpha}^{k_\rmX}(\cdot)$ and $\gM_{\omega}^{k_\rvz}(\cdot)$, the marignal distribution induced by the generator and inference models converge as $\gM_{\alpha}^{k_\rmX}p_\omega(\rmX) \rightarrow \pi_{\alpha_t}(\rmX)$ and $\gM_{\omega}^{k_\rvz}q_\phi(\rvz|\rmX) \rightarrow p_{\omega_t}(\rvz|\rmX)$, respectively. In this limit, learning the multimodal EBM via Eqn.~\ref{eq-ebm-grad} seeks to approximate the true data distribution $\pdata(\rmX)$ while simultaneously contrasting the current model against its previous state $\pi_{\alpha_t}(\rmX)$. In particular, Eqn.~\ref{eq-ebm-grad} amounts to
\begin{equation}\label{eq-ebm-grad-surrogate}
    - \gL_\pi(\alpha) \equiv \underbrace{\KL(\pdata(\rmX)||\pi_\alpha(\rmX))}_{\text{match data density}} -  \underbrace{\KL(\pi_{\alpha_t}(\rmX)||\pi_\alpha(\rmX))}_{\text{criticize itself}}
\end{equation}
This objective reflects a form of \emph{self-adversarial} learning, where the EBM is encouraged both to fit the data and to move away from its earlier approximation. At the same time, the difference of KL terms cancels the intractable partition function $\log \mZ(\alpha)$, yielding a tractable, stable EBM learning objective.

An analogous interpretation holds for the generator objective in Eqn.~\ref{eq-gen-grad}, which can be expressed as
\begin{equation}\label{eq-gen-grad-surrogate}
    - \gL_p(\omega) \equiv \underbrace{\KL(\pdata(\rmX)||p_\omega(\rmX))}_{\text{match data density}} + \underbrace{\KL(\pi_{\alpha_t}(\rmX)||p_\omega(\rmX))}_{\text{match EBM density}} +
\end{equation}
\begin{equation}
\underbrace{\E_{\pdata(\rmX)}\left[\KL(p_{\omega_t}(\rvz|\rmX)||p_{\omega}(\rvz|\rmX))\right] + \E_{\pi_{\alpha_t}(\rmX)}\left[\KL(p_{\omega_t}(\rvz|\rmX)||p_{\omega}(\rvz|\rmX))\right]}_{\text{additional surrogate KL perturbation terms}}\nonumber
\end{equation}
Thus, the shared latent generator is learned to align its marginal distribution with both the empirical data and current EBM density. The additional KL terms over the latent posterior can be viewed as a majorization step \cite{han2019divergence}. They treat $\rvz$ as part of the complete data inferred at the current iteration and provide an upper bound on the marginal likelihood, which makes optimization more tractable. Unlike prior cooperative schemes defined for single-modality data, our formulation explicitly operates on multimodal inputs. The shared latent generator is therefore encouraged to synthesize \emph{coherent} multimodal samples that are consistent across modalities and well matched to the EBM, which in turn improves EBM learning.

Learning the joint inference model by minimizing Eqn.~\ref{eq-inf-grad} matches the corresponding latent samples from multimodal real data $\rmX \sim \pdata(\rmX)$ and multimodal synthesis $\rmX \sim \pi_{\alpha_t}(\rmX)$. Specifically, given the optimal $\gM_{\alpha}^{k_\rmX}(\cdot)$ and $\gM_{\omega}^{k_\rvz}(\cdot)$ kernel Eqn.~\ref{eq-inf-grad} becomes
\begin{equation}\label{eq-inf-grad-surrogate}
    - \gL_q(\phi) \equiv \underbrace{\E_{\pdata(\rmX)}\left[\KL(p_{\omega_t}(\rvz|\rmX)||q_{\phi}(\rvz|\rmX))\right]}_{\text{real latent sample inference}} + \underbrace{\E_{ \pi_{\alpha_t}(\rmX)}\left[\KL(p_{\omega_t}(\rvz|\rmX)||q_{\phi}(\rvz|\rmX))\right]}_{\text{synthesis latent sample inference}}
\end{equation}
Both terms push the joint inference model $q\phi(\rvz|\rmX)$ toward the generator posterior $p_{\omega_t}(\rvz|\rmX)$, for both real and EBM-synthesized multimodal inputs. Since the posterior samples produced by MCMC (cf. Eqn.~\ref{eq-gen-posterior}) are typically sharper and more accurate than the initial mixture-of-experts approximation, this objective gradually improves the quality of the latent initializations. In turn, better initializations make subsequent MCMC posterior refinement more effective, allowing the generator to explore the latent space more thoroughly and to model complex multimodal structure.

\subsection{Model Generalization to Modal-specific Latent Variable}
Notably, prior works \citep{sutter2020multimodalmmJSD,palumbo2023mmvae+} extend the shared latent generative model by introducing additional modality-specific latent variables $\rmW = \{\rvw_1,\dots,\rvw_M\}$, i.e.,
\begin{equation}\label{eq-shared-gen-w}
\begin{aligned}
    p_\omega(\rmX, \rvz, \rmW)=p_0(\rvz)\prod_{i=1}^Mp_{\omega_i}(\rvx_i|\rvz, \rvw_i)p_0(\rvw_i)\;\;\;\;\;\;\;
    q_\phi(\rvz, \rmW|\rmX)=q_{\phi_\rvz}(\rvz|\rmX)\prod_{i=1}^Mq_{\phi_{\rvw_i}}(\rvw_i|\rvx_i)
\end{aligned}
\end{equation}
The modality-specific latent variable $\rmW$ is introduced to capture inductive biases unique to each modality, thereby enhancing the representational capacity of the latent space and improving robustness in cross-modal inference scenarios. Our proposed learning framework can extend to this setting. The same MCMC-revised kernels can be applied jointly over $(\rmX, \rvz, \rmW)$, so all model components are updated under the same cooperative scheme. 

\section{Related Work}\label{sec-related-work}
\noindent\textbf{Energy-Based Models.}
EBMs offer high modeling flexibility and have been widely studied in the context of maximum-likelihood training \citep{nijkamp2019learning,du2019implicit,du2020improved,xiao2020vaebm,marks2025learning,dutta2025learning}. Beyond standard MLE, recent work has explored amortized sampling using generator networks \citep{han2019divergence,grathwohl2021no,kumar2019maximum,luo2024energy}. For example, \citet{luo2024energy} proposes learning conditional EBMs, while \citet{han2019divergence,grathwohl2021no,kumar2019maximum,schroder2023energydiscrepancies} learn marginal EBMs with auxiliary or complementary generators to approximate the EBM distribution. These amortized–MCMC methods differ fundamentally from our MCMC-based training in their learning objectives (see Appendix.~\ref{sec-app-nomcmc} for discussion). Another line of work, termed cooperative learning, uses a generator to initialize MCMC chains for more efficient EBM training \citep{xie2018cooperative,xie2021learning}. However, existing methods in this family have been developed primarily for single-modal data. Extending EBMs to multimodal settings introduces additional difficulty because the sampler should also account for complex inter-modal dependencies to produce coherent, cross-modally consistent samples.

\noindent\textbf{Multimodal VAE.}
Multimodal VAEs have become a central paradigm for multimodal representation learning. MVAE \citep{wu2018multimodalmvae} introduces a product-of-experts (PoE) inference scheme that combines unimodal posteriors into a single joint posterior, enabling scalable training and principled handling of missing modalities. MMVAE \citep{shi2019variationalmmvae} instead adopts a mixture-of-experts (MoE) inference to improve robustness under partial observations. MoPoE \citep{sutter2020multimodalmmJSD,sutter2021generalizedMoPoE} unifies these ideas via a mixture-of-products formulation that interpolates between PoE and MoE, balancing flexibility and expressivity. Various recent works have also demonstrated other informative strategies for factorizing the joint posterior, such as CoDEVAE \cite{codevae}, MWBVAE \cite{mwbvae}, HELVAE \cite{helvae}, and InvariantVAE \cite{InvariantVAE}. Beyond inference structure, several works focus on refining the latent space. MVTCAE \citep{hwang2021multiMVTCAE} enforces cross-modal consistency through total-correlation regularization. MMVAE+ \citep{palumbo2023mmvae+} augments the shared latent space with modality-specific priors, and MVEBM \citep{yuan2024mvebm} replaces the Gaussian prior with an energy-based prior to capture latent structure. CMVAE \citep{palumbo2024deepcmvae} explicitly promotes semantic clustering in the shared latent space.

\noindent\textbf{Multimodal VAE with Diffusion.}
Recent multimodal methods further couple VAEs with diffusion or score-based models for refinement \citep{pandey2022,palumbo2024deepcmvae,diff-mvae,wesego2023scoremmvae}. Specifically, Diff-CMVAE \citep{palumbo2024deepcmvae} integrates DiffuseVAE \citep{pandey2022}, applying diffusion-based refinement to modality-specific outputs to significantly improve generation quality. ScoreMVAE \citep{wesego2023scoremmvae} applies score-based refinement at the latent level. These approaches demonstrate that powerful diffusion backbones can substantially boost generation quality, but they typically operate as post-hoc refinement stages. Instead, our work directly learns a multimodal EBM over the joint data space and couples it with a shared latent generator and inference model through MCMC revision, so that the core generative model itself is improved without relying on a separate refinement step.

\section{Experiment}
\noindent\textbf{Experiment Setting.} Following prior multimodal VAE work \cite{palumbo2023mmvae+,palumbo2024deepcmvae}, we benchmark our method on PolyMNIST \citep{sutter2021} and Caltech-Birds (CUB) Image–Captions \citep{shi2019variationalmmvae}. PolyMNIST contains five visual modalities that depict the same digit class under different styles and backgrounds. CUB provides paired image and text modalities with rich semantics and substantial modality-specific variation, and is regarded as a challenging benchmark for multimodal generation and alignment. Additional experiments and supplementary results are provided in Appendix.~\ref{sec-app-add-exp}.


\subsection{Multimodal Data Modelling}
\begin{minipage}[t]{\textwidth}
    \centering
    \resizebox{0.9\columnwidth}{!}{
      \begin{tabular}{c||ccccccccccc|cc}
        \toprule
        Conditional CUB  & MVAE & MVTCAE & mmJSD & MoPoE & MMVAE & MMVAE$+$ & MVEBM & MWBVAE & CoDEVAE & HELVAE & Diff-CMVAE & Ours & Ours-$\rmW$\\
        \midrule
        FID ($\downarrow$)  & 172.21 & 208.43 & 262.80 & 265.55 & 232.20 & 164.94 & 136.16 & 196.42 & 175.97 & 157.56 & 28.00 & \textbf{25.98} & \textbf{24.32}\\
        \bottomrule
    \end{tabular}
    }
\end{minipage}
\begin{minipage}{\textwidth}
    \centering
    \includegraphics[width=0.9\textwidth]{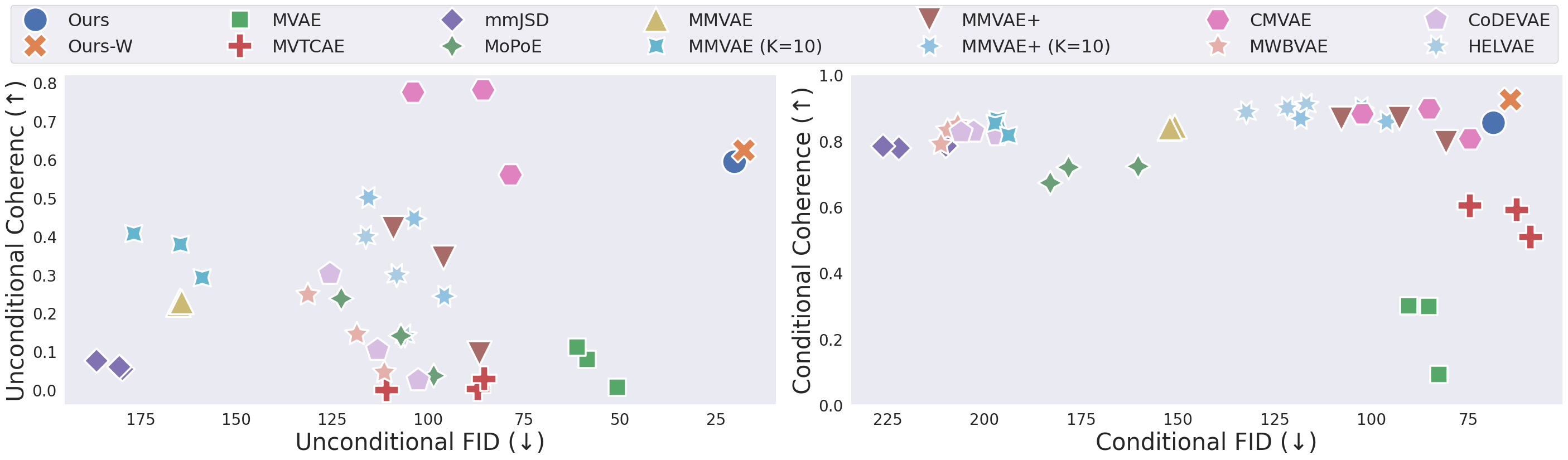}
    \captionof{figure}{Comparison for unconditional and conditional multimodal synthesis on PolyMNIST (bottom), and comparison for conditional FID score on CUB (top). Additional quantitative and qualitative results are provided in Appendix.\ref{sec-app-supp-result}.}
    \label{fig.poly-scatter}
\end{minipage}

We first assess whether our proposed learning scheme successfully yields high-quality, coherent multimodal synthesis. In our method, the generator is encouraged to approximate the EBM density and provide good initial states for EBM sampling, while the joint inference model offers informative latent initializations for posterior MCMC sampling. This cooperative interplay should contribute to improved synthesis quality and semantic coherence across modalities.

To quantitatively assess synthesis coherence, we follow standard practice and apply pre-trained classifiers\footnote{Pre-trained classifiers for each modality are provided by \citet{palumbo2024deepcmvae}.} to the generated samples. These classifiers evaluate whether generated samples correspond to the correct digit class; higher classification accuracy indicates stronger cross-modal alignment. We compare against strong multimodal VAE baselines on both unconditional and conditional generation. For CUB, the CMVAE results are taken from its diffusion-based variant (Diff-CMVAE), which integrates a diffusion model to substantially boost image quality. We denote our variant with modality-specific variables (Eqn.~\ref{eq-shared-gen-w}) as Ours-$\rmW$. 

Figure.~\ref{fig.poly-scatter} includes our variational baselines trained under different configurations (e.g., importance-weight sampling) for a broad comparison over the landscape. Across this spectrum, our method consistently achieves better coherence–quality performance, even when compared to diffusion-based methods (e.g., Diff-CMVAE \cite{palumbo2024deepcmvae}). These results demonstrate that MCMC-revised cooperative learning effectively captures shared semantics, preserves modality-specific details, and yields superior performance.

\subsection{Analysis of Learning Multimodal EBM with Complementary Models}
In this section, we study how the complementary models affect the learning of multimodal EBM.

\noindent\textbf{Learning Multimodal EBM \emph{without} Complementary Models.}
We begin with the challenging setting in which the complementary multimodal shared generator is removed. In this case, the EBM learning objective in Eqn.~\ref{eq-ebm-grad} can be viewed to minimizing only the first KL term as $\min_\alpha \KL(\Phi_{\omega_t,\phi_t}(\rmX, \rvz)||\pi_\alpha(\rmX)q_\phi(\rvz|\rmX))$, which reduces to standard MLE learning of EBM, i.e., $=\min_\alpha \KL(\pdata(\rmX)||\pi_\alpha(\rmX))$ since $\Phi_{\omega_t,\phi_t}(\rmX, \rvz)$ is anchored on the data distribution (see details in Appendix. \ref{sec-app-theo}). In this setting, we keep the same EBM architecture and perform EBM sampling using short-run Langevin dynamics initialized from random noise.

\begin{figure}[!t]
    \centering
    \setlength{\belowcaptionskip}{-15pt}%
    \includegraphics[width=0.95\textwidth]{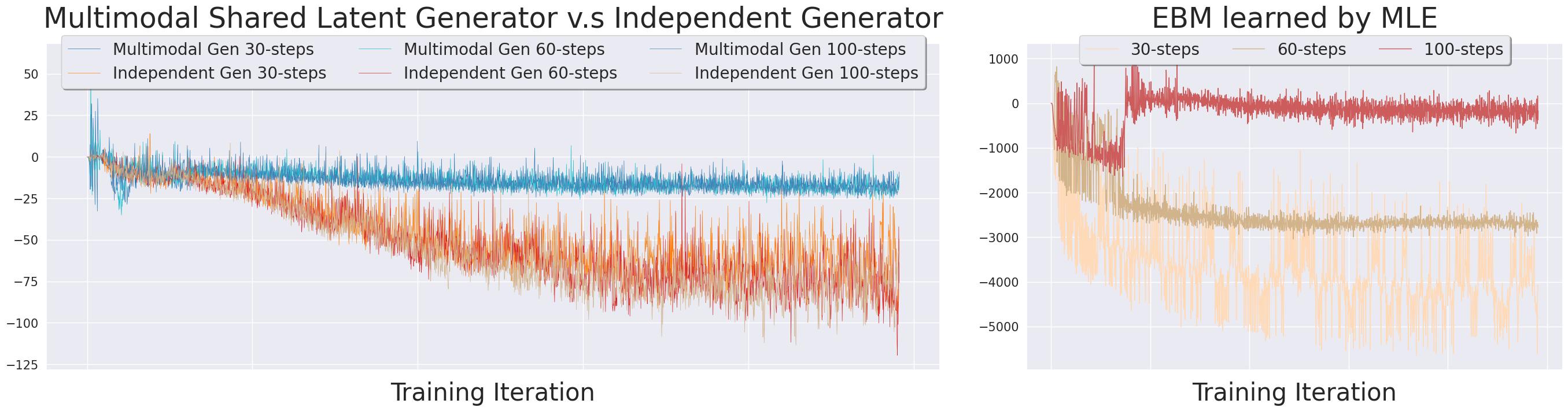}
    \caption{\textbf{Left}: EBM loss when using our shared latent generator versus independent per-modality generators; \textbf{Right}: EBM learned by MLE with noise-initialized Langevin dynamics with different sampling steps.}
    \label{fig.shared-gen}
\end{figure}

As shown in Figure.~\ref{fig.shared-gen}, the resulting EBM loss trajectory fluctuates strongly throughout training, even when we increase the number of Langevin steps. This behavior indicates that short-run MCMC from non-informative initial states does not adequately explore the high-dimensional multimodal space (see corresponding FID in Appendix. \ref{table.fid-mle-ebm}). In contrast, when we use our proposed learning framework with complementary models, the EBM loss curve becomes smoother, suggesting that the learned initializers are essential for multimodal EBM training.

\noindent\textbf{Learning Multimodal EBM with \emph{Independent} Generators.} We then examine the role of the multimodal shared latent generator. In our framework, the shared latent generator (Eqn.~\ref{eq-shared-gen}) factorizes a single latent variable $\rvz$ to capture inter-modal dependencies, thereby producing coherent multimodal initializations for EBM sampling. To assess its importance, we replace it with $M$ \emph{independent} generators $p_{\omega_i}(\rvx_i, \rvz_i)$, each with its ownlatent variable $\rvz_i$ and no explicit coupling across modalities.

Figure.~\ref{fig.shared-gen} reports the corresponding EBM loss profiles (Eqn.~\ref{eq-ebm-grad}). With independent generators, the EBM loss again shows strong fluctuations over training, even when we increase the number of EBM sampling steps (e.g., $k_{\rmX}=60$ or $100$). This behavior indicates that independently trained generators do not provide coherent multimodal initializations, making it difficult for the EBM to learn consistent cross-modal structure.
In contrast, the shared latent generator yields smoother loss curves, highlighting that a shared latent representation is crucial for generating multimodal initial states and for effective multimodal EBM learning.

\noindent\textbf{Learning Multimodal EBM with \emph{Independent} Inference Models.}

\begin{wrapfigure}{r}{0.5\textwidth}
    \centering
    \setlength{\belowcaptionskip}{-10pt}%
    \includegraphics[width=0.5\textwidth]{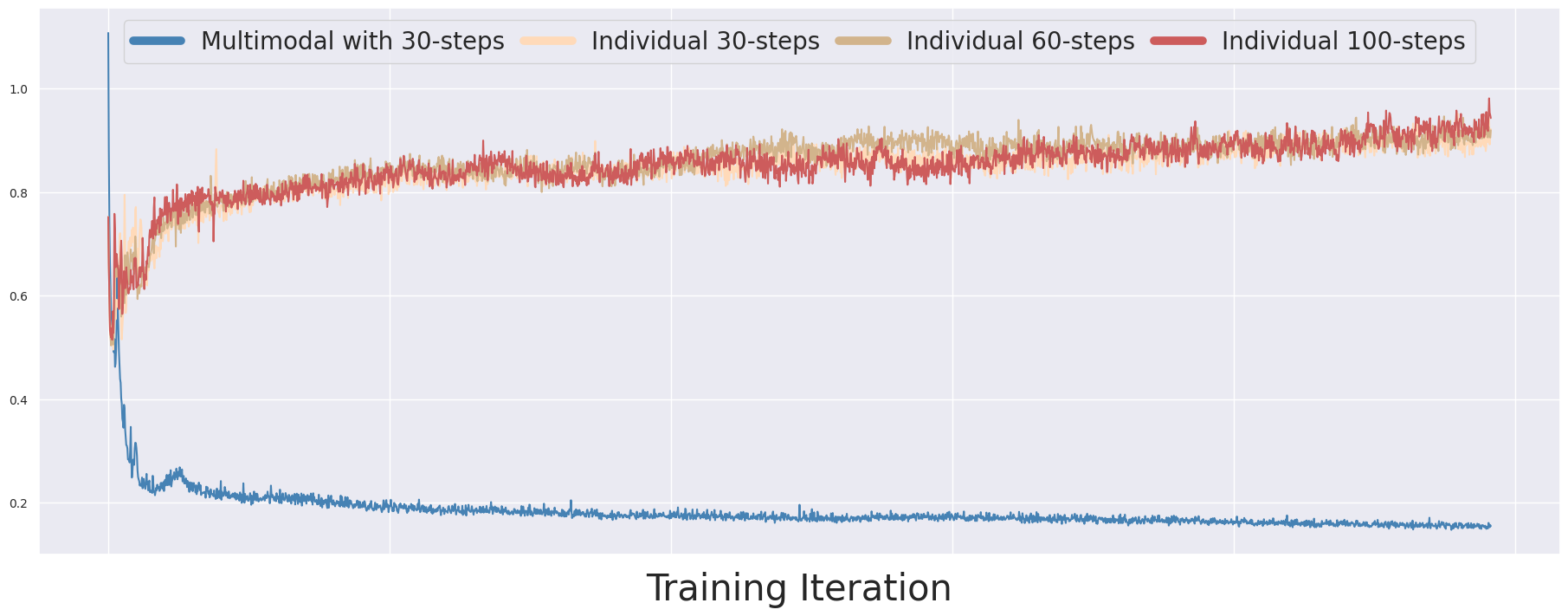}
    \caption{Generator loss profiles for joint vs. independent inference models. }
    \label{fig.joint-inf}
\end{wrapfigure}
Finally, we investigate the effect of the multimodal joint inference model, which aggregates information from all available modalities to produce a shared latent initialization for posterior MCMC. To assess its impact, we replace the joint inference model with $M$ \emph{independent} inference models $p_{\phi_i}(\rvz_i|\rvx_i)$, each predicting a separate latent variable $\rvz_i$ from a single modality and without explicit cross-modal coupling in latent space.

The corresponding generator loss profiles (Eqn.~\ref{eq-gen-grad}) are reported in Figure.~\ref{fig.joint-inf}. With independent inference models, the generator loss remains high and can even increase with more posterior sampling steps (e.g., $k_{\rvz}=60$ or $100$), indicating that the resulting latent initializations are not well aligned across modalities. In contrast, our multimodal joint inference model leads to a consistent decrease in generator loss. This suggests that shared latent initialization substantially improves generator learning, thereby supporting more effective multimodal EBM learning. 

Taken together, these three analyses highlight that (i) learning multimodal EBMs directly from noise-initialized MCMC is difficult, and (ii) both the shared latent generator and the multimodal joint inference model are important for providing coherent initial states that facilitate effective multimodal EBM learning.





\subsection{How Complementary Models Match with Their MCMC Revision?}
\begin{figure}[!h]
    \centering
    \begin{minipage}[h]{0.45\textwidth}
        \centering
        \includegraphics[width=0.92\textwidth]{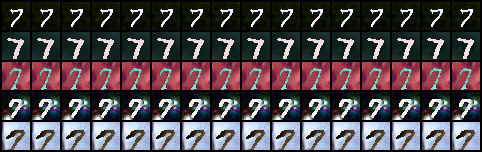}
        \includegraphics[width=0.99\textwidth]{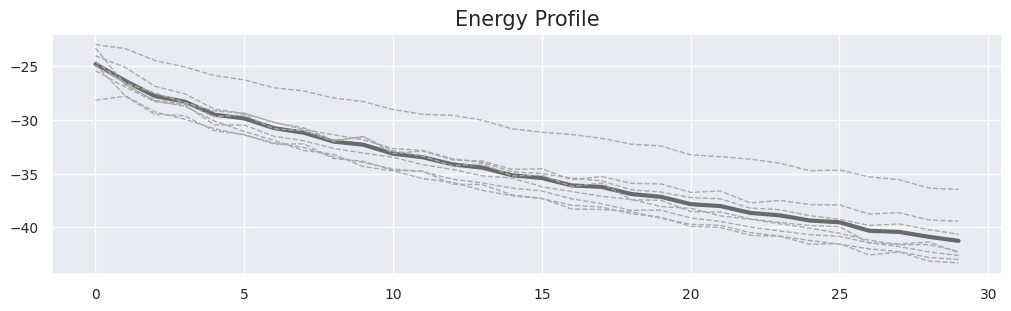}
    \end{minipage}
    \begin{minipage}[h]{0.45\textwidth}
        \centering
        \includegraphics[width=0.99\textwidth]{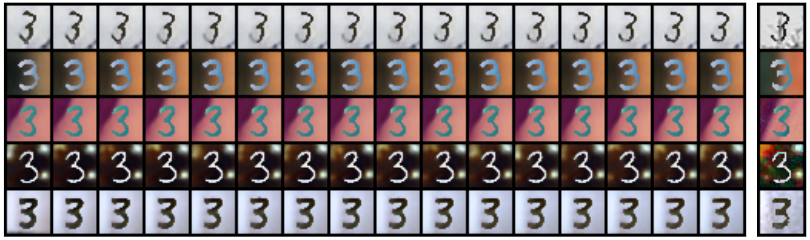}
        \includegraphics[width=0.99\textwidth]{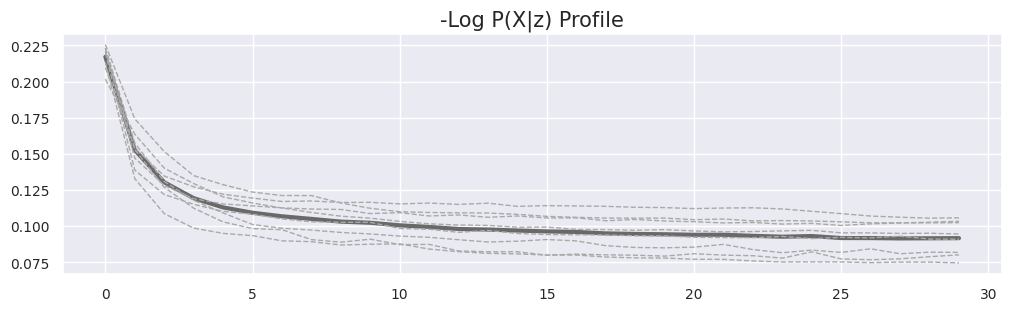}
    \end{minipage}
    \captionof{figure}{Trajectories of EBM sampling (\textbf{left}) and posterior sampling (\textbf{right}). Each row corresponds to one modality. The first column shows initial states from the initializer models; intermediate columns show intermediate samples (every 2 steps, up to 30 steps); the final column shows the MCMC-refined outputs. For posterior sampling, the rightmost column shows the observed input examples.}
    \label{fig.mcmc}
\end{figure}
It is crucial that the shared latent generator and joint inference model are well aligned with their corresponding MCMC-refined samples, so that they can provide good initial states for EBM sampling and posterior sampling, respectively. To assess this, we visualize the trajectories of EBM sampling and posterior sampling, each initialized from its complementary model and then refined via a finite number of Langevin steps.

As shown in Figure.~\ref{fig.mcmc}, the initializations from both the shared latent generator and the joint inference model are already semantically coherent across modalities (capturing the same digit class). Along the Langevin trajectories, only modest visual changes are observed, indicating that the initializer models closely match their MCMC-refined counterparts. At the same time, the energy profile $F_\alpha(\mathbf{X}^k)$ and the log-likelihood $\log p_\omega(\mathbf{X}|\mathbf{z}^k)$ continue to improve over iterations, demonstrating that the MCMC kernels still provide useful signals rather than merely reconstructing the initial states.

Overall, these observations show that the complementary models and their MCMC revisions are well matched, with the revision steps further guiding and sharpening both the EBM and the shared latent generator. In contrast to VAE-based approaches that rely on an external diffusion stage for high-quality refinement, our method learns the multimodal EBM, shared latent generator, and joint inference model in a unified framework, enabling both the EBM and the generator to produce high-quality multimodal samples directly (see additional results in Section.~\ref{sec-clip}). 

\subsection{What Does MCMC Revision Cost and Provide?}\label{sec-clip}
Our framework introduces MCMC revision, which incurs computation cost compared to purely variational generators. To make this trade-off clear, we compare sampling time, NFE (Number of Function Evaluations), and parameter overhead, together with FID and CLIP scores on CUB.

\begin{minipage}{\textwidth}
    \centering
    \captionof{table}{Comparison for sampling time (second / batch), NFE ($\downarrow$), FID ($\downarrow$), and CLIP scores ($\uparrow$).}
    \label{tab.sampling-time}
    \resizebox{0.7\columnwidth}{!}{
    \begin{tabular}{c|cc|cc|c}
    \toprule
    & Ours (Gen) & Ours (EBM) & CMVAE & Diff-CMVAE & MMVAE\\
    \midrule
     Sampling  & 0.001 & 0.08 & 0.001 & 38.12 & 0.001 \\
     NFE & 1 & 31 & 1 & 251 & 1 \\
     FID (T2I) &	26.15 & 25.98 &155.11	&28.00&	232.20\\
     FID (uncond)&	21.45 & 20.72 &	141.00&	N/A	&213.89\\
     CLIP (uncond)&	0.280 & 0.284 &	0.263&	N/A&	0.231\\
     CLIP (T2I)	&0.278 & 0.282 &	0.260&	0.272&	0.242\\
     CLIP (I2T)	&0.286 & 0.290 &	0.272&	N/A&	0.235\\
     Parameter Overhead & - &	2M	&-&	14M &	-\\
    \toprule
    \end{tabular}
    }
\end{minipage}


Table.~\ref{tab.sampling-time} shows that MCMC revision yields consistent gains in both generation quality (lower FID) and cross-modal alignment (higher CLIP) at relatively modest cost. The EBM-augmented variant improves substantially over CMVAE and MMVAE, while adding only a small number of additional parameters and keeping sampling fast. At the same time, the shared latent generator alone already achieves strong performance with essentially one-step sampling, indicating that cooperative training makes the generator itself an effective multimodal model rather than just an auxiliary sampler. In comparison, diffusion-refined VAEs (e.g., Diff-CMVAE) rely on a separately trained, second-stage diffusion model, which entails higher sampling time and a larger parameter budget. 

\subsection{Scale-up to High-resolution and Large-scale Dataset}
\begin{minipage}[h]{\textwidth}
    \centering
    \includegraphics[width=0.66\textwidth]{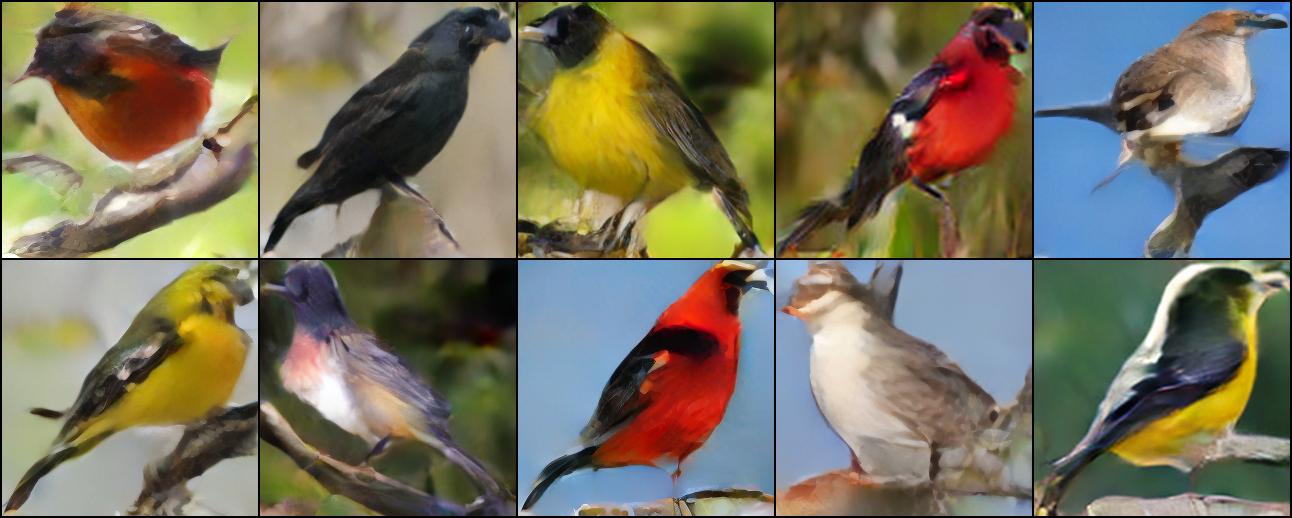}
    \includegraphics[width=0.33\textwidth]{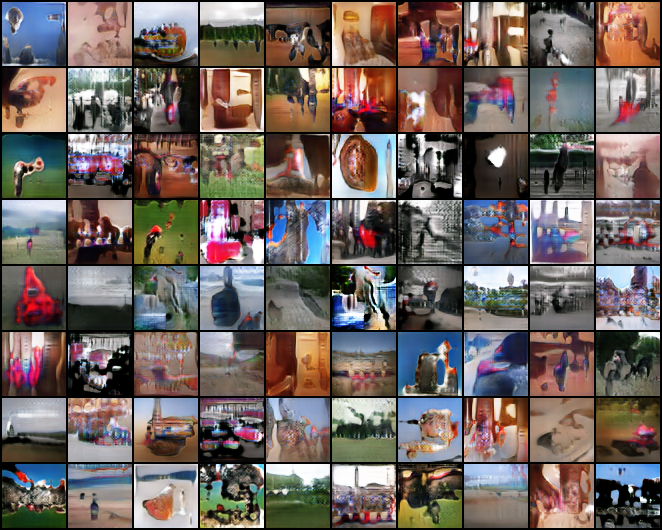}
    \captionof{figure}{Unconditional synthesis on high-resolution CUB (left) and large-scale MSCOCO (right).}
    \label{fig.scale}
\end{minipage}

\begin{wraptable}{r}{0.4\columnwidth}
    \centering
    \captionof{table}{FID ($\downarrow$) on challenging dataset.}
    \label{tab.fid-challeng}
    \resizebox{0.4\columnwidth}{!}{
    \begin{tabular}{c|ccc}
    \toprule
     & Ours (Gen) & Ours (EBM) & MMVAE+ \\
    CUB (256x256) & 56.32 & 55.81 & 213.74 \\
    MSCOCO & 68.94 & 68.10 & 187.22 \\
    \toprule
    \end{tabular}
    }
\end{wraptable}
We test the scalability of our proposed method on the challenging high-resolution image (256x256) CUB data and the large-scale MSCOCO datasets. To better understand and assess the effectiveness endowed by our proposed learning method, we use the same network structures for all experiments. We visualize the unconditional and conditional multimodal synthesis in Figure.~\ref{fig.scale}, suggesting our method effectively scales to higher resolutions and large-scale datasets while maintaining faithful multimodal synthesis quality. To further quantify this performance, we evaluate the generation quality and show results in Table.~\ref{tab.fid-challeng}.


\subsection{Ablation Study}
We investigate how the number of Langevin steps in the two MCMC kernels, $\gM_{\alpha}^{k_\rmX}$ for EBM sampling and $\gM_{\omega}^{k_\rvz}$ for posterior sampling, affects both performance and training cost (see Table.~\ref{tab.mcmcTx}).

\noindent\textbf{MCMC Steps of $\gM_{\alpha}^{k_\rmX}$.} For EBM sampling, increasing $k_\rmX$ allows Langevin dynamics to explore the energy landscape more thoroughly, which in turn provides a stronger revision signal for the shared latent generator and yields a better learned EBM. On CUB, using a small number of steps ($k_\rmX=10$) leads to noticeably worse FID but lower training time, while a larger value ($k_\rmX=60$) improves FID but also doubles the cost. We thus report the setting ($k_\rmX=30$ and $k_\rvz=30$) of a favorable balance.

\noindent\textbf{MCMC Steps of $\gM_{\omega}^{k_\rvz}$.} For posterior sampling, increasing $k_\rvz$ improves the accuracy of samples from the generator posterior, which helps the joint inference model better match $p_\omega(\rvz|\rmX)$ and thus supports more effective generator learning. When $k_\rvz$ is small (e.g., $k_\rvz=10$), FID degrades accordingly, whereas a larger value (e.g., $k_\rvz=60$) slightly improves FID but also incurs higher computational cost. The intermediate setting ($k_\rvz=30$ with $k_\rmX=30$) offers a good trade-off between performance and efficiency. These ablations suggest that MCMC revisions can yield most of the gains while keeping training practical.

\begin{minipage}{\textwidth}
    \centering
    \captionof{table}{MCMC Steps for FID and training Time.}
    \label{tab.mcmcTx}
    \resizebox{0.65\columnwidth}{!}{
    \begin{tabular}{c|ccccc}
    \toprule
     CUB & $k_\rmX$=60 & $k_\rmX$=10 & $k_\rmX$=30 and $k_\rvz$=30 & $k_\rvz$=10 & $k_\rvz$=60\\
     \midrule
     FID & 25.16 & 30.40 & \textbf{25.98} & 35.46 & 25.78  \\
     Time (seconds/iteration) & 2.62 & 1.34 & 1.78 & 1.03 & 3.15 \\
    \toprule
    \end{tabular}
    }
\end{minipage}



\section{Conclusion}
We propose a joint learning scheme that effectively learns the multimodal EBM by interweaving MLE updates of EBM, shared latent generator, and joint inference model via MCMC-based revision. The shared latent generator is learned to provide coherent initializations for EBM sampling, while the joint inference model is learned to offer starting points for posterior sampling. These MCMC samples serve as revision signals, guiding the complementary models, which in turn facilitate effective multimodal EBM sampling and learning.

\bibliography{main}
\bibliographystyle{tmlr}

\appendix
\section{Additional Experiment}\label{sec-app-add-exp}
\subsection{Latent Classification}
In our method, the joint inference model is continuously updated to catch up with the MCMC-revised posterior, which makes its latent predictions more consistent across modalities and should therefore yield a more semantically structured shared latent space. We further examine whether the inferred latent variables capture shared high-level semantics across modalities. 

\begin{minipage}{\textwidth}
    \centering
    \captionof{table}{Accuracy for latent classifier.}
    \label{tab.latent-clfs}
    \resizebox{0.5\columnwidth}{!}{
    \begin{tabular}{c|c|c|c|c}
    \toprule
     Method & Ours & MVAE & MMVAE & MoPoE \\
     Accuracy & 0.962 & 0.926 & 0.835 & 0.944 \\
    \toprule
    \end{tabular}
    }
\end{minipage}

Following prior work \citep{sutter2021generalizedMoPoE}, we train simple classifiers on the inferred latent representations and report their classification accuracy. If the shared latent space encodes meaningful semantic information, these classifiers should perform well and achieve high accuracy. Using our mixture-based joint inference model, we compute accuracy for each modality and report the average across all modalities. As shown in Table.~\ref{tab.latent-clfs}, our method attains the highest latent classification accuracy among the compared multimodal VAEs, suggesting that the learned latent space is more semantically structured and better aligned with the underlying label information.

\subsection{Latent Space Interpolation}
\begin{minipage}[h]{\textwidth}
    \centering
    \includegraphics[width=0.37\textwidth]{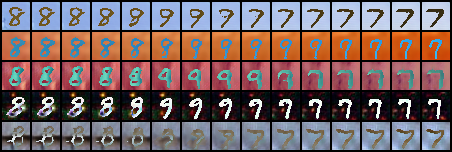}
    \includegraphics[width=0.62\textwidth]{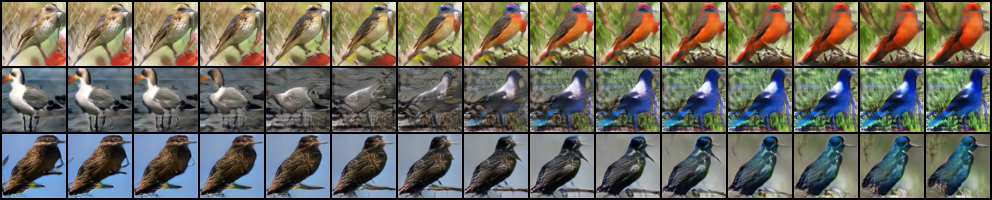}
    \captionof{figure}{Visualization of unconditional synthesis via Latent space interpolation.}
    \label{fig.interpolation}
\end{minipage}

We evaluate whether the shared latent generator model can produce smooth interpolations in the shared latent space, leading to gradual transitions in the multimodal data space. To this end, we perform linear interpolation in the latent space, $\tilde{\rvz} = (1-\alpha)\cdot\rvz_1 + \alpha\cdot\rvz_2$. As shown in Figure.~\ref{fig.interpolation}, the shared latent generator produces smooth and coherent transitions across modalities, indicating its ability to capture shared semantics and effectively explore the energy landscape.

\subsection{MCMC Refinement on Cross-modal Inference}
Following the evaluation protocols of our variational baselines, we assessed conditional coherence when only one modality is available in Figure.~\ref{fig.poly-scatter} in the main text. In our framework, MCMC posterior sampling offers an additional capability: it can refine latent variables inferred from multiple subsets of available modalities. This refinement step, which iteratively adjusts the latent variables toward better cross-modal consistency, is, however, not feasible for standard variational approaches.

In particular, given a set of observed modality data, our proposed method performs MCMC refinement in the latent space to further improve inference quality. Formally, let $\rmX_\gO$ denote the set of observed modalities indexed by $\gO \in \{1, \cdots, M\}$. The generator posterior conditioned on this partial observation can be defined as $p_\omega(\rvz|\rmX_\gO) \propto p_0(\rvz)\prod_{m \in \gO}p_{\omega_m}(\rvx_m|\rvz)$. The corresponding Langevin posterior refinement uses the gradient 
\begin{equation}
    \frac{\partial}{\partial \rvz}\log p_\omega(\rvz|\rmX_\gO)=\frac{\partial}{\partial \rvz}\log p_0(\rvz)+\sum_{m \in \gO}\frac{\partial}{\partial \rvz}\log p_{\omega_m}(\rvx_m|\rvz)
\end{equation} 
This subset-conditioned posterior follows from the factorization of the multimodal shared latent generative model and is defined under partial observations. We note that our multimodal generator is learned using full multimodal data, which learns a shared latent representation that captures inter-modal semantic structure.

\begin{minipage}{\textwidth}
    \centering
    \captionof{table}{Coherence with MCMC refinement.}
    \label{tab.mcmc-refinement}
    \resizebox{0.65\columnwidth}{!}{
    \begin{tabular}{c|c|c|c|c}
    \toprule
     Number of Modality & 1 ($n=0$) & 2 ($n=1$) & 3 ($n=2$) & 4 ($n=3$) \\
     Coherence & 0.921 & 0.930 & 0.938 & 0.940 \\
    \toprule
    \end{tabular}
    }
\end{minipage}

If our model is well-learned, the role of Langevin refinement is to improve the latent representation by incorporating evidence from the additional available modalities. Starting from an initialization $\rvz_0 \sim q_{\phi}(\rvz|\rmX_\gO)$, the refinement step adjusts the latent variable toward regions that better explain the observed subset under the generator likelihood. In this way, the refinement process seeks a latent representation that more closely corresponds to a shared multimodal mode consistent with the available observations. We report our results in Table.~\ref{tab.mcmc-refinement}, where the conditional coherence becomes better with increasing number of available modalities under our MCMC refinement process.

\section{Theorectical Derivation}\label{sec-app-theo}
For MLE learning of the EBM objective (Eqn.~\ref{eq-ebm-mle-grad}), the gradient is derived as
\begin{equation}\label{eq-app-ebm-mle-grad}
\begin{aligned}
        \frac{\partial}{\partial \alpha}\gL_\pi(\alpha) = \E_{\pdata(\rmX)}\left[\frac{\partial}{\partial \alpha}\log \pi_\alpha(\rmX)\right]
        = \E_{\pdata(\rmX)}\left[\frac{\partial}{\partial \alpha}F_\alpha(\rmX)\right] - \frac{\partial}{\partial \alpha}\log \mZ(\alpha)
\end{aligned}
\end{equation}
where $\frac{\partial}{\partial \alpha}\log \mZ(\alpha)$ is derived as
\begin{equation}\label{eq-app-ebm-mle-grad-Z}
\begin{aligned}
\frac{\partial}{\partial \alpha}\log \mZ(\alpha) = \frac{1}{\mZ(\alpha)}\int\frac{\partial}{\partial \alpha}\exp \left[F_\alpha(\rmX)\right]\rvd\rmX
=\int\pi_\alpha(\rmX)\frac{\partial}{\partial \alpha}\left[F_\alpha(\rmX)\right]\rvd\rmX
=\E_{\pi_\alpha(\rmX)}\left[\frac{\partial}{\partial \alpha}F_\alpha(\rmX)\right]
\end{aligned}
\end{equation}
By applying Eqn.~\ref{eq-app-ebm-mle-grad-Z} to Eqn.~\ref{eq-app-ebm-mle-grad}, we have derived Eqn.~\ref{eq-ebm-mle-grad}.

For our EBM learning objective in Eqn. \ref{eq-ebm-grad}, the gradient is derived from the KL difference formulation
\begin{equation}\label{eq-app-ebm-Z-cancel}
    - \hat{\gL}_\pi(\alpha) = \KL(\Phi_{\omega_t,\phi_t}(\rmX, \rvz)||\pi_\alpha(\rmX)q_\phi(\rvz|\rmX)) -  \KL(\Omega_{\omega_t, \alpha_t}(\rmX,\rvz)||\pi_\alpha(\rmX)q_\phi(\rvz|\rmX))
\end{equation}
\begin{equation}
\begin{aligned}
\text{with}\;\;\; \frac{\partial}{\partial \alpha}\hat{\gL}_\pi(\alpha) 
    &= \underbrace{\E_{\Phi_{\omega_t, \alpha_t}(\rmX,\rvz)}\left[\frac{\partial}{\partial \alpha}F_\alpha(\rmX)\right] - \frac{\partial}{\partial \alpha}\log \mZ(\alpha)}_{\text{First KL Term}} - \underbrace{\E_{\Omega_{\omega_t, \alpha_t}(\rmX,\rvz)}\left[\frac{\partial}{\partial \alpha}F_\alpha(\rmX)\right] + \frac{\partial}{\partial \alpha}\log \mZ(\alpha)}_{\text{Second KL Term}}\nonumber\\
    &= \E_{\Phi_{\omega_t, \alpha_t}(\rmX,\rvz)}\left[\frac{\partial}{\partial \alpha}F_\alpha(\rmX)\right] - \E_{\Omega_{\omega_t, \alpha_t}(\rmX,\rvz)}\left[\frac{\partial}{\partial \alpha}F_\alpha(\rmX)\right] \nonumber
\end{aligned}
\end{equation}
in which, with respect to the gradient on $\alpha$, the first KL divergence term corresponds to the MLE learning of EBM with samples anchored by the data distribution through $\Phi_{\omega_t, \alpha_t}(\rmX,\rvz)=\pdata(\rmX)\cdot\gM_{\omega_t}^{k_\rvz}\cdot q_{\phi_t}(\rvz|\rmX)$. For the second KL-divergence of its own critique, the partition function is canceled out, and samples are obtained from the MCMC-revised distribution $\Omega_{\omega_t, \alpha_t}(\rmX,\rvz)=\gM_{\alpha_t}^{k_\rmX}\cdot p_{\omega_t}(\rmX|\rvz)p_0(\rvz)$ at the current optimization step under the stop-gradient update scheme.

For MLE learning of the shared latent generator objective (Eqn.~\ref{eq-gen-mle-grad}), the gradient is derived as
\begin{equation}\label{mle-grad}
\begin{aligned} 
    &\frac{\partial}{\partial \omega}\log p_{\omega}(\rmX) = \E_{p_\omega(\rvz|\rmX)}[\frac{\partial}{\partial \omega}\log p_\omega(\rmX)]\\
    &=\E_{p_\omega(\rvz|\rmX)}[\frac{\partial}{\partial \omega}\log p_\omega(\rmX)] + \E_{p_\omega(\rvz|\rmX)}[\frac{\partial}{\partial \omega}\log p_\omega(\rvz|\rmX)]\\
    &=\E_{p_\omega(\rvz|\rmX)}[\frac{\partial}{\partial \omega}\log p_\omega(\rmX, \rvz)]
\end{aligned}
\end{equation}
where $\E_{p_\omega(\rvz|\rmX)}[\frac{\partial}{\partial \omega}\log p_\omega(\rvz|\rmX)]=\int p_\omega(\rvz|\rmX)[\frac{\partial}{\partial \omega}\log p_\omega(\rvz|\rmX)] \rvd\rvz=\frac{\partial}{\partial \omega}\int p_\omega(\rvz|\rmX)\rvd\rvz=0$.

\subsection{Compared to Amorized-MCMC Method}\label{sec-app-nomcmc}
Several recent advances have investigated EBM learning without explicit MCMC sampling \cite{grathwohl2021no,han2019divergence,luo2024energy,schroder2023energy}. These works study \textit{single-modal} EBMs that employ amortized samplers to replace MCMC, thereby avoiding iterative sampling. In contrast, our focus is on the \textit{multimodal setting}, which introduces two additional challenges: (i) effectively capturing the shared inter-modal relationships across heterogeneous modalities, and (ii) mitigating the mismatch induced by multimodal joint inference models. To address these challenges, we incorporate MCMC revision as a key component of our cooperative framework, which allows both the EBM and the generator posterior to be iteratively refined by each other, ensuring coherent multimodal alignment that cannot be achieved by amortized single-pass updates.

Moreover, the inclusion of MCMC revision makes our \textbf{learning objectives fundamentally different} from previous amortizing formulations. For clarity, and to directly illustrate the difference in learning dynamics independent of modality notation, we denote $\Omega,\Phi$ as shorthand for MCMC-revised densities $\Omega_{\omega,\alpha}(\mathbf{X}, \mathbf{z}), \Phi_{\omega,\phi}(\mathbf{X},\mathbf{z})$  and denote $Q=q_\phi(\mathbf{z}|\mathbf{X})p_{data}(\mathbf{X})$, $\Pi=\pi_\alpha(\mathbf{X})q_\phi(\mathbf{z}|\mathbf{X})$,$P=p_\omega(\mathbf{X},\mathbf{z})$ for joint densities of amortized models. We denote AM for methods using amortized models without MCMC.

\textbf{Learning the EBM ($\alpha$):} The corresponding KL terms in our method (Eqn.~\ref{eq-ebm-grad}) and AM are: $\min_\alpha KL(\Phi\| \Pi) - KL(\Omega\|\Pi)$ v.s. $\min_\alpha KL(Q\| \Pi) - KL(P\|\Pi)$. Our formulation leverages MCMC-revised samples (i.e., joint densities of $\Omega$ and $\Phi$), whereas AM relies solely on ancestral samples ($Q$ and $P$). Because our samples are refined by the EBM itself, they provide a more accurate approximation of the target energy landscape $KL(M_{\alpha_t}p_{\omega_t}(\mathbf{X})\|\pi_{\alpha_t}(\mathbf{X}))\le KL(p_{\omega_t}(\mathbf{X})\|\pi_{\alpha_t}(\mathbf{X}))$. This results in more effective and stable EBM learning and leverages the contextual modelling capability of EBM to effectively guide the multimodal generator model.

\textbf{Learning the (shared) generator model ($\omega$):} For learning the generator model (Eqn. 11), KL terms for ours and AM are $\min_\theta KL(\Phi\|P)+KL(\Omega\| P)$ v.s. $\min_\theta KL(Q\|P) + KL(P\| \Pi)$. The learning dynamics differ substantially. In our case, the generator is trained with MCMC-revised latent samples, yielding a closer match to the true generator posterior: $KL(M_{\omega_t}q_{\phi_t}(\mathbf{z}|\mathbf{X})\|p_{\omega_t}(\mathbf{z}|\mathbf{X}))\le KL(q_{\phi_t}(\mathbf{z}|\mathbf{X})\|p_{\theta_t}(\mathbf{z}|\mathbf{X}))$, which aims to address the mismatch between the generaetor posterior and joint inference model (analysis in Sec. 3.1). In addition, $KL(\Omega\|P)$ learns to match the revised MCMC samples from EBM-refined samples, while the Amortizer method intends to chase the major modes of $\pi_\alpha(\mathbf{X})$ through variational approximation (i.e., $KL(P\| \Pi)$). Hence, our generator directly learns from revised multimodal samples that can better capture inter-modal consistency.

\textbf{Learning the (joint) inference model ($\phi$):} For the inference model (Eqn. 12), learning objectives for ours and AM are: $\min_\phi KL(\Phi\| Q) + KL(\Omega\|\Pi)$ v.s. $\min_\phi KL(Q\|P) + KL(P\|\Pi)$. The two approaches differ in both learning source and optimization target. Our inference network amortizes latent MCMC refinement on observed data (i.e., $KL(\Phi\| Q)ComplilerError$), while AM performs pure variational inference (i.e., $KL(Q|P)$). On generated samples, our model matches EBM-revised generator samples (i.e., $KL(\Omega|\Pi)$), whereas AM directly uses ancestral generator outputs ($KL(P|\Pi)$), which can lead to sub-optimal inference quality.

\section{Inference Mechanism under Missing Modalities}
Our framework follows the standard inference mechanism established in multimodal VAEs for handling missing modalities. Given any available observed modality $\mathbf{x}_i$, we first infer its shared latent variable, and then use this latent variable to generate the missing modalities through the shared latent generator for $\mathbf{x}_j$ where $j \ne i$. This mechanism is identical to prior multimodal VAE baselines, ensuring fair comparison and consistent inference behavior. In our experiments, we evaluate using the same inference mechanism as in baseline models to ensure fairness, and the results consistently show superior reconstruction and coherence. 

\section{Implementation}\label{sec-app-implement}
\begin{minipage}[h]{\textwidth}
    \begin{minipage}{0.49\textwidth}
    \centering
    \resizebox{0.99\columnwidth}{!}{
        \begin{tabular}{|c|}
        \hline 
            \textbf{EBM Block} (in\_ch, out\_ch, downsample, head)\\
        \hline  
            Input: x \\
            ReLU if head \\
            Conv(in\_ch, out\_ch), ReLU, Conv(out\_ch, out\_ch)\\
            Downsample(factor=2) if downsample\\
            output: h\\
            \hline
            Input: x \\
            Downsample(factor=2) if head\\
            Conv(in\_ch, out\_ch) if downsample\\
            Downsample(factor=2) if downsample and not head\\   
            output: y\\
            \hline
            output: h + y\\
        \hline
        \end{tabular}
    }
    \end{minipage}
    \begin{minipage}{0.49\textwidth}
    \centering
    \resizebox{0.99\columnwidth}{!}{
        \begin{tabular}{|c|}
        \hline 
            \textbf{EBM Network on PolyMNIST} (nef)\\
        \hline  
            Input: $\rmX$ \\
            h = concat(Conv(each $\rvx$)) along channel dim\\
            EBM Block(nc, nef, downsample=True, head=True)\\
            EBM Block (nef, nef, downsample=True)\\
            EBM Block (nef, nef, downsample=False)\\
            EBM Block (nef, nef, downsample=False)\\
            ReLU, Downsample(factor=8), Linear(nef, 1)\\
            output: h\\
        \hline
            \textbf{EBM Network on CUB} (nef)\\
        \hline  
            Input: $\rmX$ \\
            img = Conv(img), txt=Linear(ReLu(Linear(txt emb)))\\
            EBM Block(nc, nef, downsample=True, head=True)\\
            EBM Block (nef, nef, downsample=True)\\
            EBM Block (nef, nef, downsample=False)\\
            EBM Block (nef, nef, downsample=False)\\
            ReLU, Downsample(factor=8), Linear(Concat(h,txt), 1)\\
            output: h\\
        \hline
        \end{tabular}
    }
    \end{minipage}
    \captionof{table}{We use generator and inference network structures from \citep{palumbo2023mmvae+, palumbo2024deepcmvae}. For our EBM energy function structures, we denote the operation of convolution as \textbf{Conv} (input channel, output channel, k=3, s=1, p=1), where $\rm k$ is the kernel size, $\rm s$ is the stride number, and $\rm p$ is padding value. We conduct Upsample and Downsample via \textit{interpolate} and \textit{avg\_pool2d} operations.}
    \label{table.networks}
\end{minipage}

\section{Supplementary Result}\label{sec-app-supp-result}
Corresponding to our Figure.~\ref{fig.poly-scatter}, we additionally report quantitative results in Table.~\ref{table.coherence} and Table.~\ref{table.fid}, where we report only the best performance of our baselines. Qualitative results can be seen in Figure.~\ref{fig.cub_unconditional_syn}, Figure.~\ref{fig.cub_conditional_syn}, Figure.~\ref{fig.poly_unconditional_syn}, and Figure.~\ref{fig.poly_cross}.

\begin{center}
\begin{minipage}[h]{\textwidth}
    \begin{minipage}{0.45\textwidth}
        \centering
        \captionof{table}{Comparison of synthesis coherence.}
        \label{table.coherence}
        \resizebox{0.925\columnwidth}{!}{
          \begin{tabular}{c||cc}
            \toprule
            \multirow{2}{*}{Methods}  & \multicolumn{2}{c}{PolyMNIST}\\
            & Unconditional & Conditional\\
            \toprule
            MVAE       & 0.112  & 0.301  \\
            MVTCAE     & 0.029  & 0.604  \\
            mmJSD      & 0.076  & 0.785  \\
            MoPoE      & 0.238  & 0.723  \\ 
            MMVAE      & 0.232  & 0.844  \\
            MMVAE$+$   & 0.421  & 0.869  \\
            MVEBM      & 0.735  & 0.857  \\
            CMVAE      & 0.781  & 0.897  \\
            MWBVAE     & 0.297  & 0.850 \\
            CoDEVAE    & 0.431  & 0.828  \\
            HELVAE     & 0.508  & 0.910  \\
            \midrule
            \textbf{Ours} & \textbf{0.594} & \textbf{0.855}\\
            \textbf{Ours-$\rmW$} & \textbf{0.624} & \textbf{0.921}\\
            \toprule
        \end{tabular}
        }
    \end{minipage}
    \hfill
    \begin{minipage}{0.55\textwidth}
        \centering
        \captionof{table}{Comparison of multimodal synthesis quality. }
        \label{table.fid}
        \resizebox{0.76\columnwidth}{!}{
          \begin{tabular}{c||cc}
            \toprule
            \multirow{2}{*}{Methods}  & \multicolumn{2}{c}{PolyMNIST}\\
            & Unconditional & Conditional\\
            \toprule
            MVAE       & 50.65  & 82.59 \\
            MVTCAE     & 85.43  & 58.95 \\
            mmJSD      & 179.76 & 178.27 \\
            MoPoE      & 98.56  & 160.29 \\ 
            MMVAE      & 164.29 & 150.83 \\
            MMVAE$+$   & 86.64  & 80.75 \\
            MVEBM      & 75.43  & 70.45 \\
            CMVAE      & 78.52  & 74.53  \\
            MWBVAE     & 111.45 & 206.88 \\
            CoDEVAE    & 102.61 & 196.64 \\
            HELVAE     & 106.05 & 116.83 \\
            \midrule
            \textbf{Ours} & \textbf{20.12} & \textbf{68.52}\\
            \textbf{Ours-$\rmW$} & \textbf{17.65} & \textbf{64.12}\\
            \toprule
        \end{tabular}
        }
    \end{minipage}
\end{minipage}    
\end{center}

\begin{center}
    \begin{minipage}[h]{\textwidth}
        \centering
        \captionof{table}{Unconditional FID on EBM learned by MLE.}
        \label{table.fid-mle-ebm}
        \resizebox{0.425\columnwidth}{!}{
          \begin{tabular}{c||ccc}
            \toprule
            \multirow{2}{*}{Ours}  & \multicolumn{3}{c}{MLE learned}\\
            & 30-steps & 60-steps & 100-steps\\
            \toprule
            20.12 & 125.23 & 101.34 & 122.92 \\
            \toprule
            \toprule
        \end{tabular}
        }
\end{minipage}
\end{center}

\begin{center}
\begin{minipage}[h]{0.9\textwidth}
    \centering
    \includegraphics[width=0.99\textwidth]{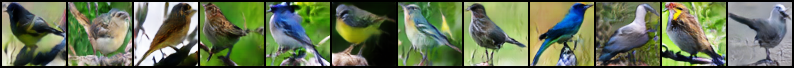}
    \resizebox{0.9\columnwidth}{!}{
    \begin{tabular}{l}
    \toprule
     1. A black bird is up with a short, short bill. \\
     2. The bird has a small surface and ooak tree which are black yellowed branches.\\
     3. This bird has yellow with brown on its chest and has a very short beak.\\
     4. This bird has wings that are black and have a brown crown.\\
     5. This is a blue bird bird with white chest.\\
     6. The bird has a green chest and black eye rings.\\
     7. This particular bird has a belly that has white and yellow color.\\
     8. The bird has a small brown bill with brown shoulder that also appear to be juvenile.\\
     9. A blue bird with a chevron and something.\\
     10. This bird has a white neck and wings that are grey and has a short bill.\\
     11. This bird is brown coloured with a redhead and has a long crest.\\
     12. This bird is white and grey in color, with it having few black wings.\\
    \toprule
    \end{tabular}
    }
    \captionof{figure}{Unconditional generation on CUB.}
    \label{fig.cub_unconditional_syn}
\end{minipage}
\end{center}
\begin{center}
\begin{minipage}[h]{0.99\textwidth}
    \centering
    \resizebox{0.7\columnwidth}{!}{
    \begin{tabular}{l}
    \toprule
     Input: this bird is shiny black, and blue in color, with a black beak. \\
    \toprule
    \end{tabular}
    }
    \includegraphics[width=0.9\textwidth]{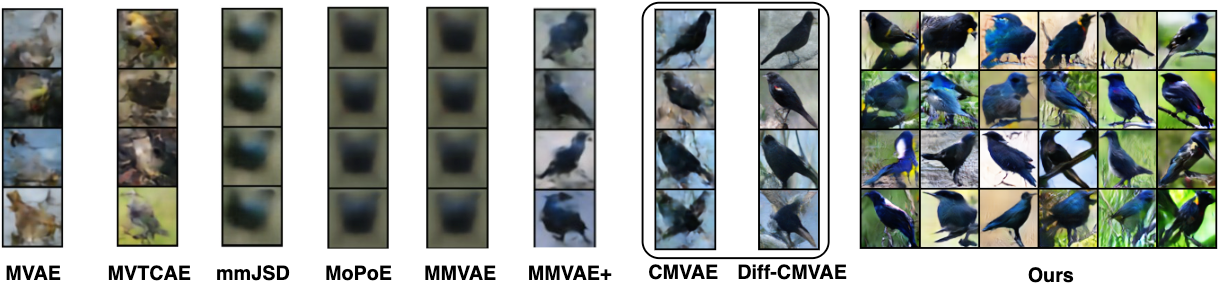}
    \captionof{figure}{Conditional generation on CUB. Baseline results are taken from \citep{palumbo2023mmvae+}. CMVAE and Diff-CMVAE results are reproduced with codes provided by \cite{palumbo2024deepcmvae,pandey2022diffusevae}.}
    \label{fig.cub_conditional_syn}
\end{minipage}
\end{center}
\begin{center}
\begin{minipage}[h]{0.9\textwidth}
    \centering
    \includegraphics[width=0.19\textwidth]{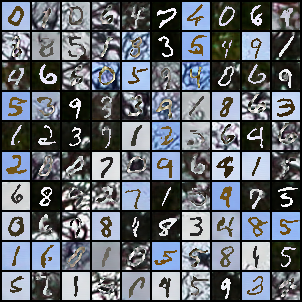}
    \includegraphics[width=0.19\textwidth]{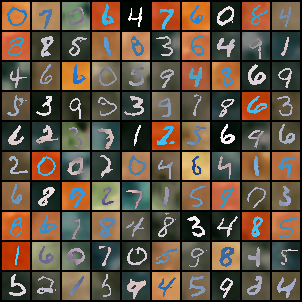}
    \includegraphics[width=0.19\textwidth]{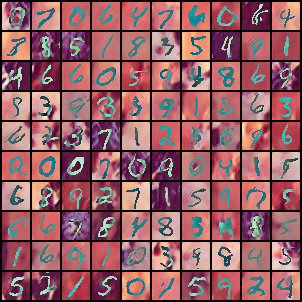}
    \includegraphics[width=0.19\textwidth]{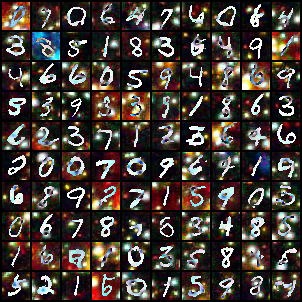}
    \includegraphics[width=0.19\textwidth]{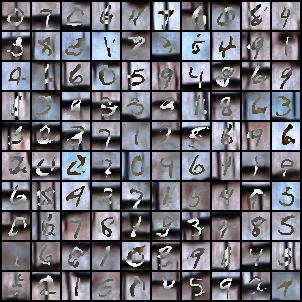}
    \captionof{figure}{Unconditional generation on PolyMNIST.}
    \label{fig.poly_unconditional_syn}
\end{minipage}
\end{center}

\begin{center}
\begin{minipage}[h]{0.9\textwidth}
    \begin{minipage}{0.99\textwidth}
        \centering
        \includegraphics[width=0.09\textwidth]{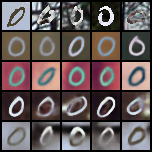}
        \includegraphics[width=0.09\textwidth]{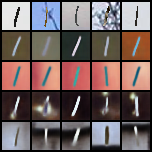}
        \includegraphics[width=0.09\textwidth]{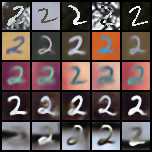}
        \includegraphics[width=0.09\textwidth]{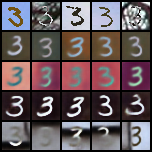}
        \includegraphics[width=0.09\textwidth]{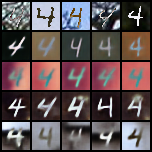}
        \includegraphics[width=0.09\textwidth]{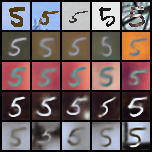}
        \includegraphics[width=0.09\textwidth]{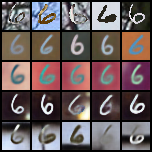}
        \includegraphics[width=0.09\textwidth]{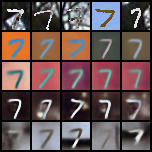}
        \includegraphics[width=0.09\textwidth]{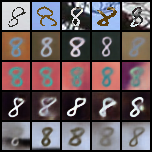}
        \includegraphics[width=0.09\textwidth]{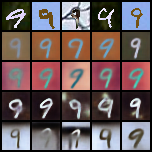}
    \end{minipage}
    \begin{minipage}{0.99\textwidth}
        \centering
        \includegraphics[width=0.09\textwidth]{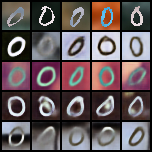}
        \includegraphics[width=0.09\textwidth]{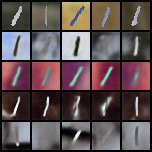}
        \includegraphics[width=0.09\textwidth]{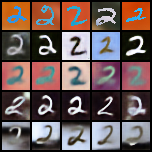}
        \includegraphics[width=0.09\textwidth]{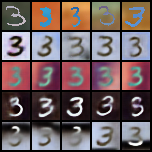}
        \includegraphics[width=0.09\textwidth]{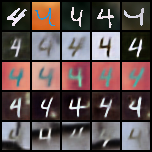}
        \includegraphics[width=0.09\textwidth]{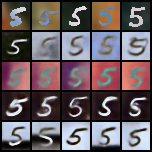}
        \includegraphics[width=0.09\textwidth]{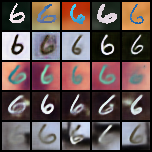}
        \includegraphics[width=0.09\textwidth]{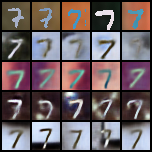}
        \includegraphics[width=0.09\textwidth]{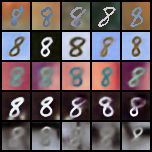}
        \includegraphics[width=0.09\textwidth]{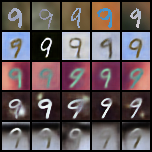}
    \end{minipage}

    \begin{minipage}{0.99\textwidth}
        \centering
        \includegraphics[width=0.09\textwidth]{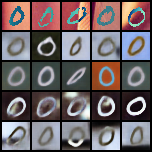}
        \includegraphics[width=0.09\textwidth]{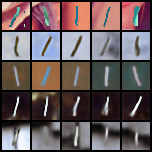}
        \includegraphics[width=0.09\textwidth]{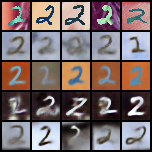}
        \includegraphics[width=0.09\textwidth]{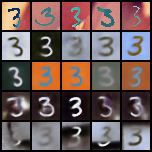}
        \includegraphics[width=0.09\textwidth]{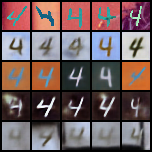}
        \includegraphics[width=0.09\textwidth]{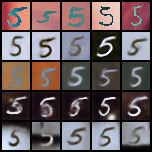}
        \includegraphics[width=0.09\textwidth]{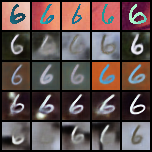}
        \includegraphics[width=0.09\textwidth]{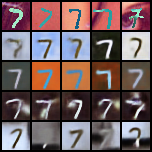}
        \includegraphics[width=0.09\textwidth]{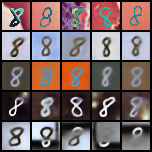}
        \includegraphics[width=0.09\textwidth]{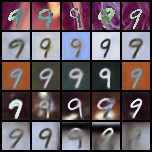}
    \end{minipage}
    \hfill
    \begin{minipage}{0.99\textwidth}
        \centering
        \includegraphics[width=0.09\textwidth]{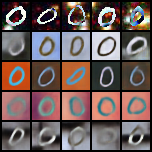}
        \includegraphics[width=0.09\textwidth]{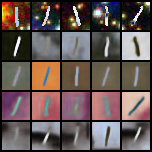}
        \includegraphics[width=0.09\textwidth]{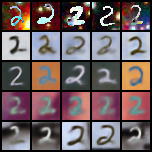}
        \includegraphics[width=0.09\textwidth]{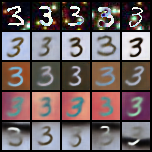}
        \includegraphics[width=0.09\textwidth]{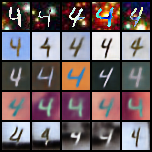}
        \includegraphics[width=0.09\textwidth]{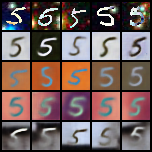}
        \includegraphics[width=0.09\textwidth]{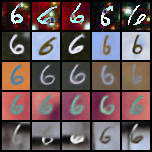}
        \includegraphics[width=0.09\textwidth]{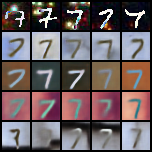}
        \includegraphics[width=0.09\textwidth]{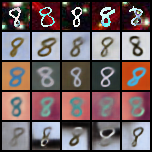}
        \includegraphics[width=0.09\textwidth]{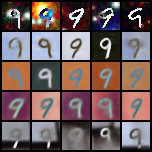}
    \end{minipage}
    \hfill
    \begin{minipage}{0.99\textwidth}
        \centering
        \includegraphics[width=0.09\textwidth]{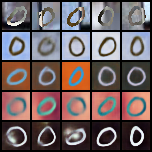}
        \includegraphics[width=0.09\textwidth]{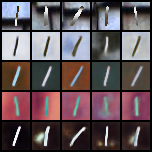}
        \includegraphics[width=0.09\textwidth]{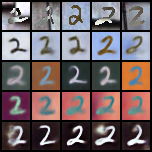}
        \includegraphics[width=0.09\textwidth]{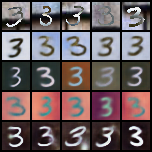}
        \includegraphics[width=0.09\textwidth]{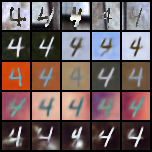}
        \includegraphics[width=0.09\textwidth]{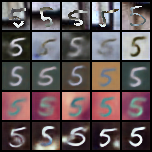}
        \includegraphics[width=0.09\textwidth]{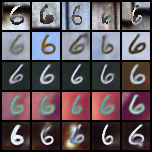}
        \includegraphics[width=0.09\textwidth]{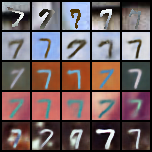}
        \includegraphics[width=0.09\textwidth]{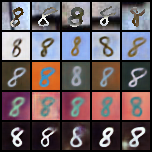}
        \includegraphics[width=0.09\textwidth]{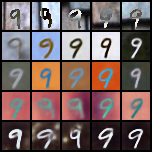}
    \end{minipage}
    \captionof{figure}{Conditional generation on PolyMNIST. From top to bottom, available modality from 1 to 5. In each block, the first row shows the given input modality, while the subsequent rows display the generated outputs for the remaining missing modalities.}
    \label{fig.poly_cross}
\end{minipage}
\end{center}

\section{Broader Impact Statement}
This work advances multimodal energy-based generative modeling, and therefore shares the usual risks associated with powerful generative models, such as potential misuse for creating misleading synthetic content or amplifying biases present in training data. At the same time, better joint modeling of multimodal structure may benefit applications such as modality prediction, provided that practitioners use appropriate safeguards and responsible data practices.

\section{Algorithm}
\begin{minipage}{0.99\textwidth}
    \begin{algorithm}[H]
    \caption{Learning Scheme}
    \label{alg:learning}
    \begin{algorithmic}[1]
	\Require
	a multimodal EBM $\alpha$, a multimodal shared latent generator $\omega$, a joint inference model $\phi$, Markov transition kernels with stop-gradient for both $\hat{\gM}_{\alpha}^{k_\rmX}(\cdot)$ and $\hat{\gM}_{\omega}^{k_\rvz}(\cdot)$ with Langevin steps $k_\rmX$ and $k_\rvz$;
    \For{$\#\,\text{training iteration}$}
      \State Draw training sample $\rmX_{\text{data}} \sim \pdata(\rmX)$.
      \State Draw latent prior sample $\rvz_{\text{prior}} \sim p_0(\rvz)$.
      \Statex \hspace{\algorithmicindent} \textcolor{gray}{$\triangleright$ \textit{MCMC-revised sample}.}
      \State Draw inference latent sample $\rvz_{\text{Inf}} \sim q_\phi(\rvz|\rmX_{\text{data}})$ with $\rmX_{\text{data}}$.
      \State Draw generator revised latent sample $\rvz_{\text{Gen-revised}} \sim \hat{\gM}_{\omega}^{k_\rvz}(\cdot)$ by Eqn.~\ref{eq-gen-ld} based on $\rvz_{\text{Inf}}$.
      \State Draw generator sample $\rmX_{\text{Gen}} \sim p_\omega(\rmX|\rvz_{\text{prior}} )$ with prior latent sample $\rvz_{\text{prior}}$.
      \State Draw EBM revised sample $\rmX_{\text{EBM-revised}} \sim \hat{\gM}_{\alpha}^{k_\rmX}(\cdot)$ by Eqn.~\ref{eq-ebm-ld} based on $\rmX_{\text{Gen}}$.
      \State Draw EBM revised latent sample $\rvz_{\text{EBM-revised}} \sim q_\phi(\rvz|\rmX_{\text{EBM-revised}})$ based on $\rmX_{\text{EBM-revised}}$.
      \Statex \hspace{\algorithmicindent} \textcolor{gray}{$\triangleright$ \textit{Model update}.}
      \State Update EBM parameter $\alpha$ using Eqn. \ref{eq-ebm-grad} with $\rmX_{\text{data}}$ and $\rmX_{\text{EBM-revised}}$.
      \State Update Generator parameter $\omega$ using Eqn. \ref{eq-gen-grad} with $\rmX_{\text{data}}$, $\rvz_{\text{Gen-revised}}$, $\rmX_{\text{Gen}}$, and $\rmX_{\text{EBM-revised}}$.
      \State Update Inference parameter $\phi$ using Eqn. \ref{eq-inf-grad} with $\rvz_{\text{Inf}}$, $\rvz_{\text{Gen-revised}}$, $\rvz_{\text{prior}}$, and $\rvz_{\text{EBM-revised}}$.
    \EndFor
    \end{algorithmic}
    \end{algorithm}
\end{minipage}

\section{Pytorch PseudoCode}\label{sec-app-code}
\begin{lstlisting}[style=pystyle,caption={PyTorch code used in our experiments.},label={lst:pytorch}]
import torch as t
import torch.nn as nn

data_loader = get_dataloader(dataset, batch_size)
netG, netI, netE = get_networks(dataset)

optG = t.optim.Adam(netG.parameters(), lr=1e-3)
optE = t.optim.Adam(netE.parameters(), lr=4e-4)
optI = t.optim.Adam(netI.parameters(), lr=1e-3)

e_l_steps, e_l_step_size, e_n_step_size = 30, 0.1, 0.001
z_l_steps, z_l_step_size, z_n_step_size = 30, 0.1, 0.1

latent_dim = 32
pz = get_distribution(t.distributions.Normal, latent_dim)
qz = get_distribution(t.distributions.Normal, latent_dim)

dataset = "PolyMNIST"
batch_size = 256

mse = nn.MSELoss(reduction='none').cuda()

def log_mean_exp(value, dim=0, keepdim=False):
    return t.logsumexp(value, dim, keepdim=keepdim) - math.log(value.size(dim))

def langevin_x(x_init):
    x = [x.clone().detach().requires_grad(True) for x in x_init]

    for steps in range(e_l_steps):
        energy = netE(x)
        energy = energy.sum()
        grad = t.autograd.grad(energy, x)
        for d, x_i in enumerate(x):
            x_i.data = x_i.data - 0.5 * e_l_step_size * e_l_step_size * grad[d] + e_n_step_size * t.randn_like(x_i).data

    return [x_i.detach() for x_i in x]

def langevin_z(z_init, x_data):
    z = [z.clone().detach().requires_grad(True) for z in z_init]
    views = len(z_init)
    for steps in range(z_l_steps):
        recon_value = [[None for _ in range(views)] for _ in range(views)]

        for e in range(views):
            for d in range(views):
                rec = netG(z[e], v_idx=d)
                recon_value[e][d] = mse(rec, x_data[d])

        nls = []
        for r in range(views):
            lpz = pz.log_prob(z[r])
            nlpx = [px_u for px_u in recon_value[r]]
            nlpxu = t.stack(nlpxu).sum(0)
            nl = nlpxu - lpz
            nls.append(nlw)
        nls = t.stack(nls).mean(0)
        nls = nls.sum(0)

        grad = t.autograd.grad(nls, z)
        for d, z_i in enumerate(z):
            z_i.data = z_i.data - 0.5 * z_l_step_size * z_l_step_size * grad[d] + z_n_step_size * t.randn_like(z_i).data

    return [z_i.detach() for z_i in z]

for i, x in enumerate(data_loader):
    x = [x_i.cuda() for x_i in x]
    views = len(x)
    
    z_prior = pz.rsample()
    samples_init = netG(z_prior)
    samples_corr = langevin_x(samples_init)

    z_q_mu_init, z_q_lv_init = netI(x)
    z_q_init = qz(z_q_mu_init, z_q_lv_init)
    z_q_corr = langevin_z(z_q_init, x)

    optG.zero_grad()
    recon_value = [[None for _ in range(views)] for _ in range(views)]

    for e in range(views):
        for d in range(views):
            rec = netG(z[e], v_idx=d)
            recon_value[e][d] = mse(rec, x[d])

    nls = []
    for r in range(views):
        nlpx = [px_u for px_u in recon_value[r]]
        nlpxu = t.stack(nlpxu).sum(0)
        nls.append(nlpxu)
    nls = t.stack(nls).mean(0)
    nls = nls.mean(0)

    errS = mse(samples_init, samples_corr)
    errG = nls + errS
    errG.backward()
    optG.step()

    optI.zero_grad()
    z_p_mu, z_p_lv = netI(samples_corr)

    nlqz_true = []
    nlqz_fake = []
    for r in range(views):
        lqz_true = log_mean_exp(t.stack([sum_flat(qz_x.log_prob(z_q_corr[r])) for qz_x in qz(z_q_mu_init, z_q_lv_init)]))
        nlqz_true.append(- lqz_true)
        lqz_gen = log_mean_exp(t.stack([sum_flat(qz_x.log_prob(z_prior)) for qz_x in qz(z_p_mu, z_p_lv)]))
        nlqz_fake.append(- lqz_gen)

    nlqz_true = t.stack(nlqz_true).mean(0)
    nlqz_fake = t.stack(nlqz_fake).mean(0)
    errI = nlqz_true + nlqz_fake
    errI.backward()
    optI.step()

    optE.zero_grad()
    E_t = netE(x)
    E_f = netE(samples_corr)
    errE = (E_t - E_f) / (e_n_step_size/e_l_step_size)**2
    errE.backward()
    optE.step()    
\end{lstlisting}

\end{document}